\documentclass[10pt,twocolumn,letterpaper]{article}

\usepackage{cvpr}              %

\usepackage{graphicx}
\usepackage{amsmath}
\usepackage{amssymb}
\usepackage{booktabs}

\usepackage[square,numbers,sort&compress]{natbib}
\usepackage{enumitem}
\usepackage{algpseudocode}
\usepackage[dvipsnames]{xcolor}
\usepackage[font=small,labelfont=bf]{caption}
\usepackage{array}
\usepackage{multirow}
\usepackage{booktabs}
\usepackage{algorithm}
\usepackage{subcaption}
\usepackage[normalem]{ulem}
\usepackage{xparse}
\usepackage{pifont}
\usepackage{bm}
\usepackage{threeparttable}
\usepackage{etoolbox}

\usepackage{listings}
\usepackage{mwe} %
\usepackage{makecell}
\usepackage{color, colortbl}
\usepackage{tabularx}
\usepackage{pifont}%
\usepackage[accsupp]{axessibility}

\usepackage[scaled=0.85]{DejaVuSansMono}

\makeatletter
\@namedef{ver@everyshi.sty}{}
\makeatother
\usepackage{tikz}
\usepackage{pgfplots}
\pgfplotsset{compat=1.16}

\definecolor{citecolor}{HTML}{0071bc}
\usepackage[breaklinks=true,bookmarks=false,colorlinks,bookmarks=false, citecolor=citecolor]{hyperref}

\usepackage[capitalize]{cleveref}
\crefname{section}{\S}{\S\S}
\crefname{subsection}{\S}{\S\S}
\crefformat{table}{Table~#2#1#3}
\crefformat{figure}{Figure~#2#1#3}
\crefformat{equation}{Eq~#2#1#3}

\newlength\savewidth

\newlength\thinwidth

\definecolor{Gray}{gray}{0.92}
\definecolor{DarkGray}{gray}{0.5}
\newcolumntype{a}{>{\columncolor{Gray}}c}
\newcolumntype{H}{>{\setbox0=\hbox\bgroup}c<{\egroup}@{}}
\definecolor{LightCyan}{rgb}{0.88,1,1}
\definecolor{altRowColor}{gray}{0.92}
\definecolor{highlightRowColor}{rgb}{0.95, 0.95, 1}

\newcommand{\demph}[1]{\textcolor{DarkGray}{#1}}

\definecolor{GrayNumber}{gray}{0.5}
\definecolor{GrayXMark}{gray}{0.7}
\newcommand{\xmark}{ {\color{GrayXMark} \ding{55}} } %

\definecolor{ImageDark}{rgb}{0,0.3,0.8}
\definecolor{VideoDark}{rgb}{.5,.0,.5}
\definecolor{ThreeDDark}{rgb}{0,.5,0}
\colorlet{Image}{ImageDark!20!white}
\colorlet{Video}{VideoDark!20!white}
\colorlet{ThreeD}{ThreeDDark!20!white}
\colorlet{ImageLight}{ImageDark!70!white}
\colorlet{VideoLight}{VideoDark!70!white}
\colorlet{ThreeDLight}{ThreeDDark!70!white}

\newcolumntype{i}{>{\columncolor{Image}}c}
\newcolumntype{v}{>{\columncolor{Video}}c}
\newcolumntype{t}{>{\columncolor{ThreeD}}c}
\newcolumntype{I}{>{\columncolor{ImageLight}}c}
\newcolumntype{V}{>{\columncolor{VideoLight}}c}
\newcolumntype{T}{>{\columncolor{ThreeDLight}}c}

\newcommand{\OURS}{\textsc{Omnivore}\xspace}

\newcommand{\swinB}{Swin-B\xspace}
\newcommand{\swinS}{Swin-S\xspace}
\newcommand{\swinT}{Swin-T\xspace}
\newcommand{\swinL}{Swin-L\xspace}

\newcommand{\imnet}{ImageNet-1K\xspace}
\newcommand{\imnetShort}{IN1K\xspace}
\newcommand{\imnetFull}{ImageNet-21K\xspace}
\newcommand{\imnetFullShort}{IN21K\xspace}
\newcommand{\kinetics}{Kinetics-400\xspace}
\newcommand{\kineticsShort}{K400\xspace}

\newcommand{\placesThree}{Places-365\xspace}
\newcommand{\placesThreeShort}{P365\xspace}
\newcommand{\pets}{Oxford-IIIT Pets\xspace}
\newcommand{\petsShort}{Pets\xspace}
\newcommand{\inat}{iNaturalist-2018\xspace}
\newcommand{\inatShort}{iNat18\xspace}
\newcommand{\sunrgbd}{SUN RGB-D\xspace}
\newcommand{\sunrgbdShort}{SUN\xspace}

\newcommand{\nyu}{NYU-v2\xspace}
\newcommand{\nyuShort}{NYU\xspace}
\newcommand{\nyuSeg}{NYU-v2-seg\xspace}
\newcommand{\nyuSegShort}{NYU-seg\xspace}

\newcommand{\sthsth}{Something Something-v2\xspace}
\newcommand{\sthsthShort}{SSv2\xspace}
\newcommand{\epic}{EPIC-Kitchens-100\xspace}
\newcommand{\epicShort}{EK100\xspace}

\newcommand{\app}{\raise.17ex\hbox{$\scriptstyle\sim$}}
\newcommand{\bI}{\mathbf{I}}
\newcommand{\bV}{\mathbf{V}}
\newcommand{\bD}{\mathbf{D}}

\newcommand{\bX}{\mathbf{X}}

\newcommand{\bx}{\mathbf{x}}
\newcommand{\be}{\mathbf{e}}

\newcommand{\bPhi}{\mathbf{\Phi}}

\newcommand{\sota}{state-of-the-art\xspace}

\def\cvprPaperID{***} %
\def\confName{CVPR}
\def\confYear{2022}

\title{\OURS: A Single Model for Many Visual Modalities}

\author{
  \resizebox{\linewidth}{!}{
    \begin{tabular}{c}
      Rohit Girdhar$^{*}$ \quad
      Mannat Singh$^*$ \quad
      Nikhila Ravi$^*$ \quad
      Laurens van der Maaten \quad
      Armand Joulin \quad
      Ishan Misra$^{*}$ \\
      {Meta AI \vspace{-0.03in}} \\
    \end{tabular}%
  } \\
  {\small \url{https://facebookresearch.github.io/omnivore}}
}

\begin{document}

\twocolumn[{%
  \renewcommand\twocolumn[1][]{#1}%
  \maketitle
  \begin{center}
    \centering
    \captionsetup{type=figure}
    \resizebox{\linewidth}{!}{
    \begin{tabular}{c@{\hskip 0.1in}|c@{\hskip 0.1in}|c@{\hskip 0.1in}|c}
        {\Large \textbf{Image}} (RGB) & {\Large \textbf{Depth map}} (D) & {\Large \textbf{Single-view 3D}} (RGBD) & {\Large \textbf{Video}} (RGBT) \\
        \includegraphics[height=0.15\linewidth]{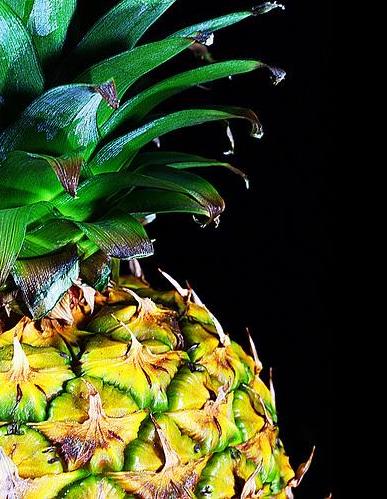} &
        \includegraphics[height=0.15\linewidth]{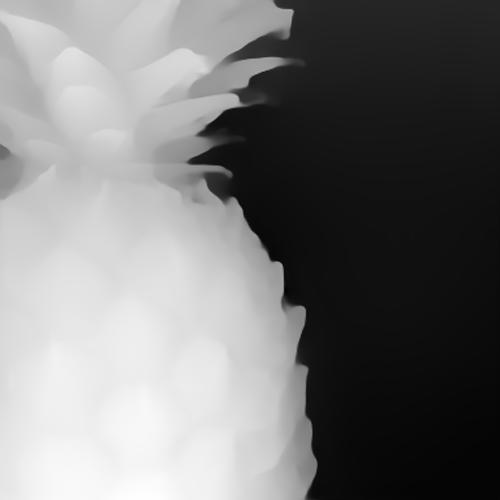} \hfill
        \includegraphics[height=0.15\linewidth]{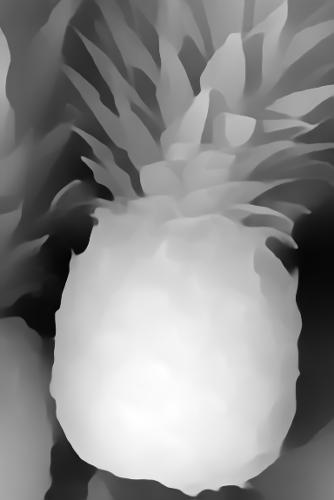} \hfill
        \includegraphics[height=0.15\linewidth]{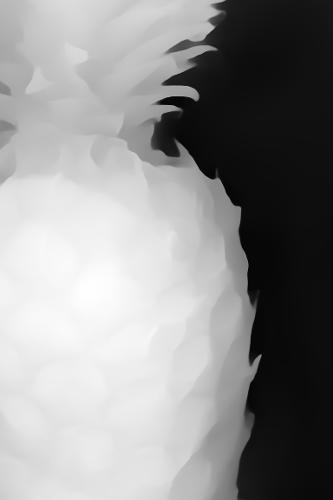}
        \hfill
        \includegraphics[height=0.15\linewidth]{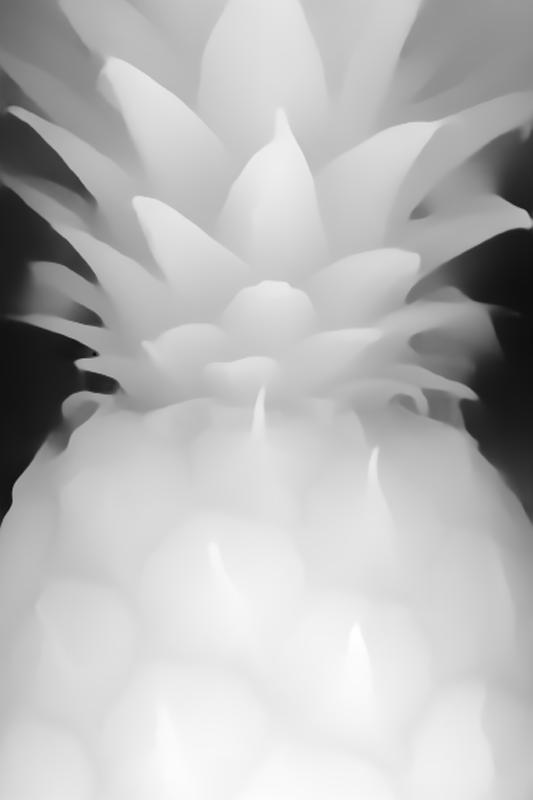}
        \hfill
        \includegraphics[height=0.15\linewidth]{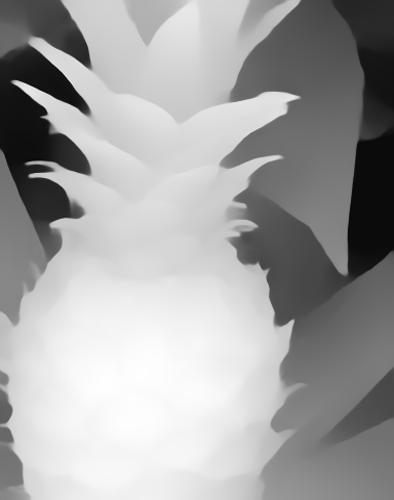}
        &
        \includegraphics[height=0.15\linewidth]{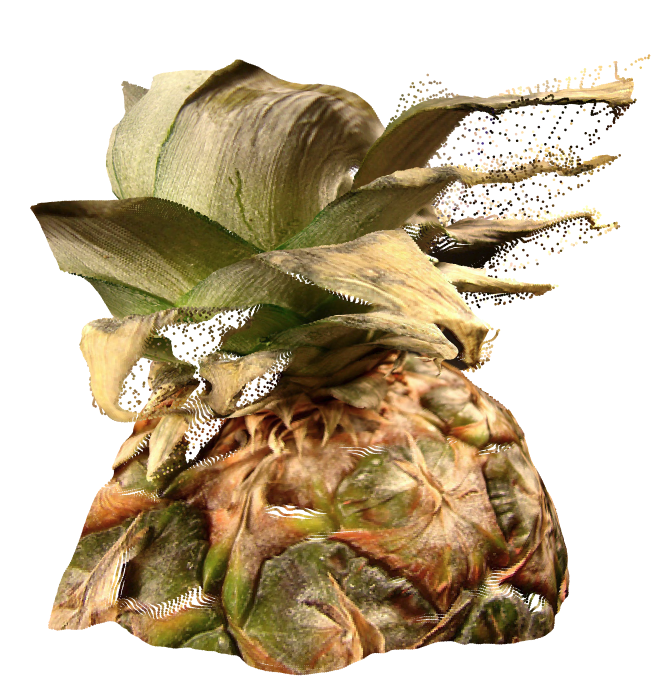} \hfill
        \includegraphics[height=0.15\linewidth]{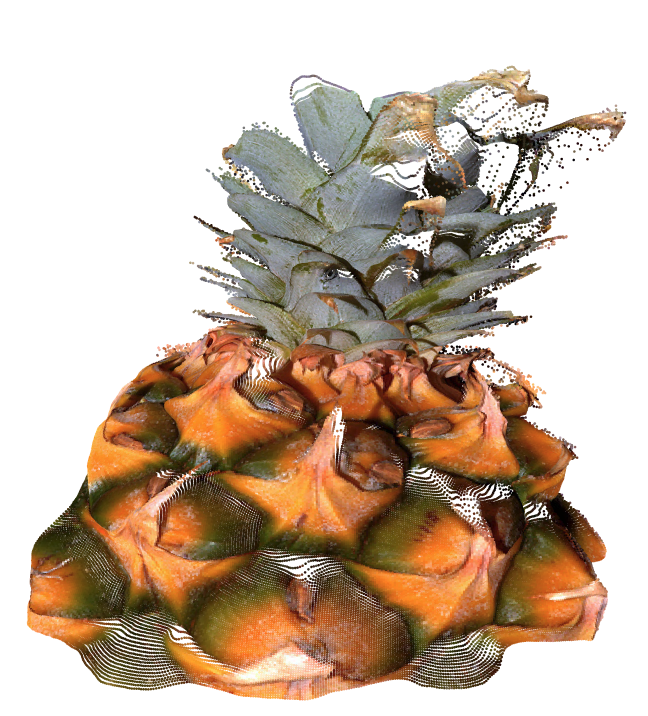} \hfill
        \includegraphics[height=0.15\linewidth]{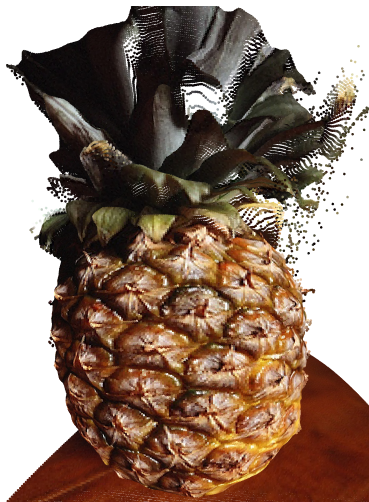}
        &
        \includegraphics[height=0.15\linewidth]{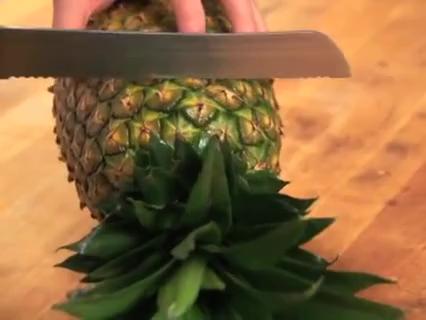} \hfill
        \includegraphics[height=0.15\linewidth]{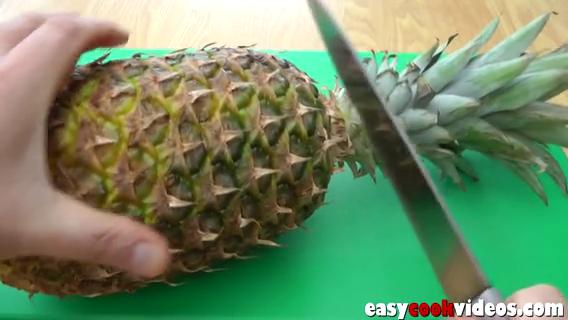}
        \\
        \includegraphics[height=0.15\linewidth]{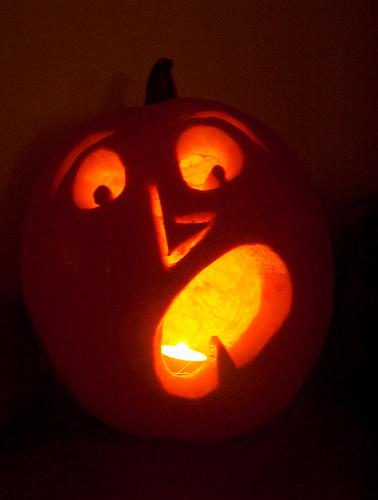} &
        \includegraphics[height=0.15\linewidth]{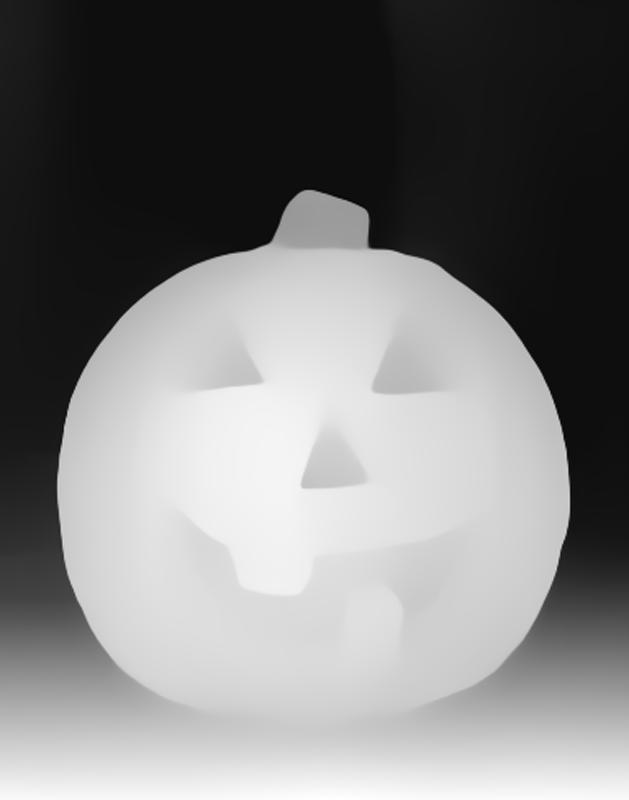} \hfill
        \includegraphics[height=0.15\linewidth]{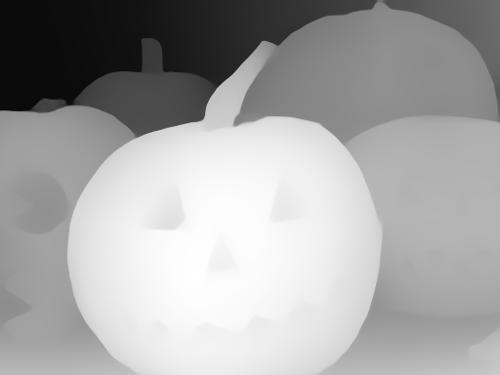} \hfill
        \includegraphics[height=0.15\linewidth]{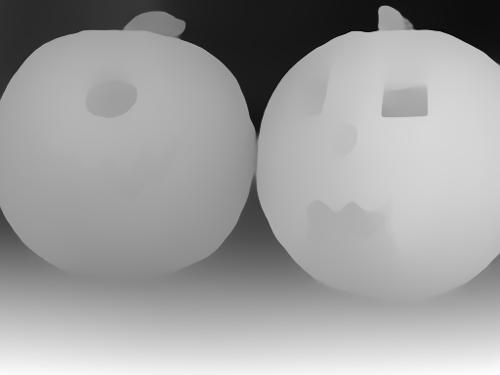} \hfill
        \includegraphics[height=0.15\linewidth]{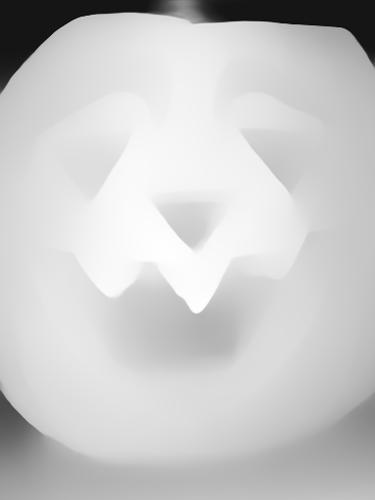}
        &
        \includegraphics[height=0.15\linewidth]{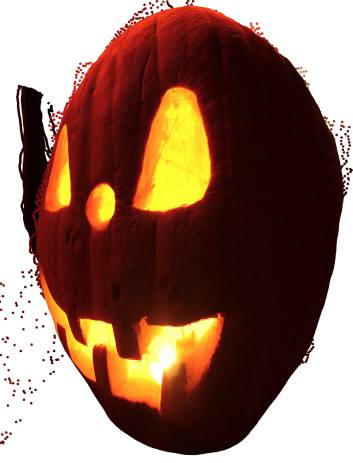} \hfill
        \includegraphics[height=0.15\linewidth]{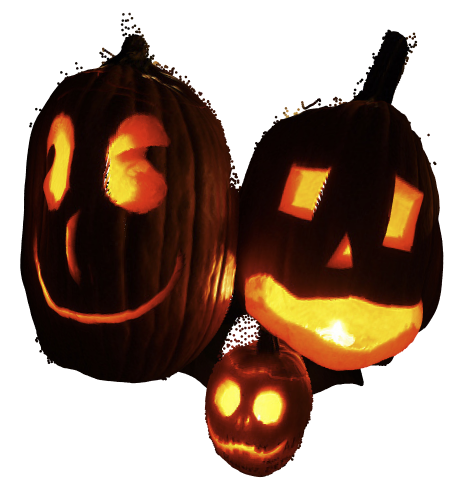} \hfill
        \includegraphics[height=0.15\linewidth]{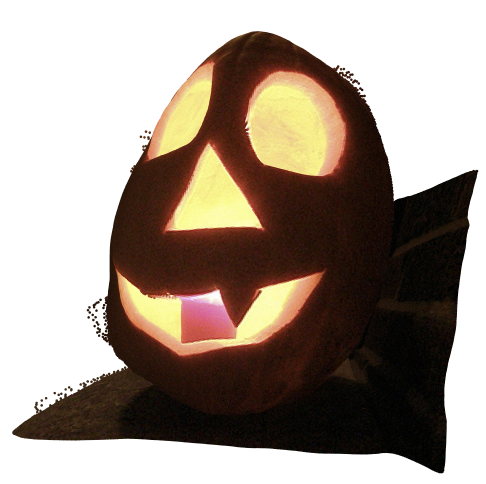} &
        \includegraphics[height=0.15\linewidth]{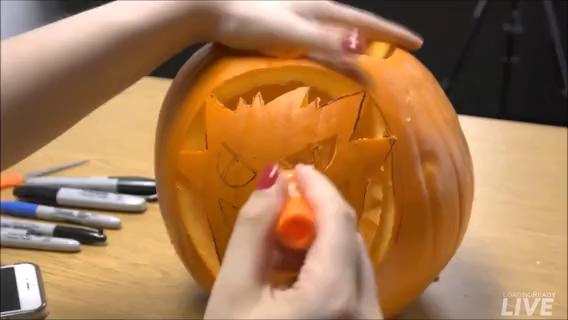} \hfill
        \includegraphics[height=0.15\linewidth]{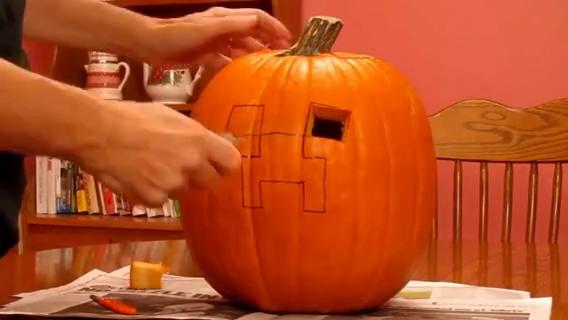}
        \\
        \includegraphics[height=0.15\linewidth]{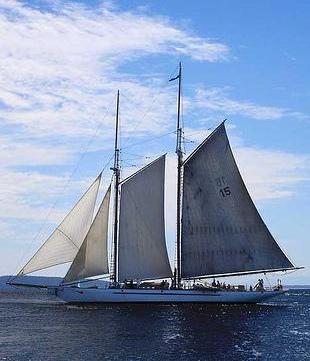} &
        \includegraphics[height=0.15\linewidth]{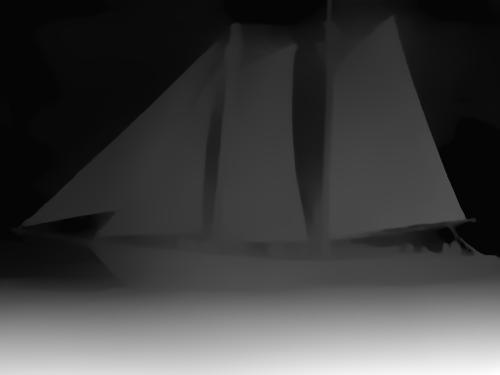} \hfill
        \includegraphics[height=0.15\linewidth]{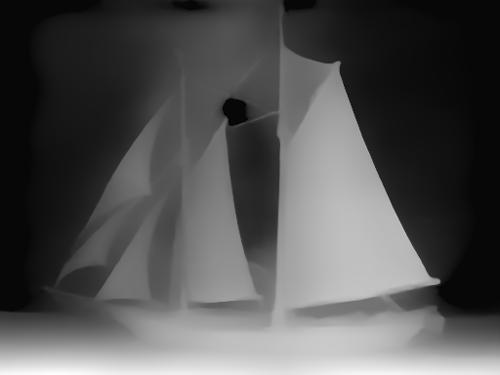} \hfill
        \includegraphics[height=0.15\linewidth]{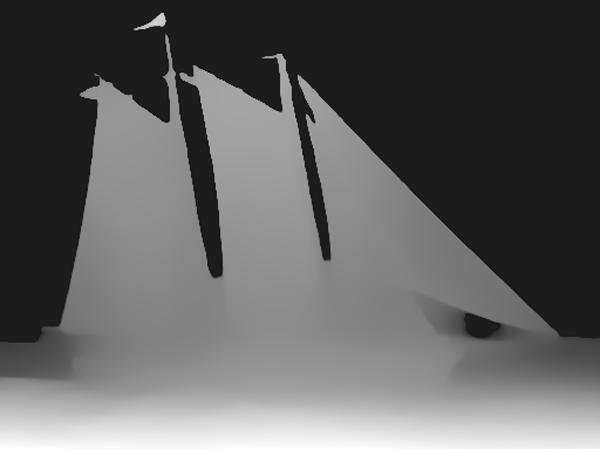} &
        \includegraphics[height=0.15\linewidth]{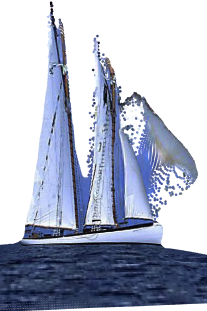} \hfill
        \includegraphics[height=0.15\linewidth]{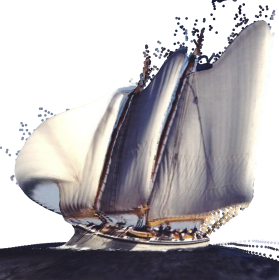} \hfill
        \includegraphics[height=0.15\linewidth]{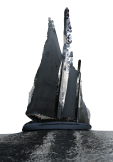} &
        \includegraphics[height=0.15\linewidth]{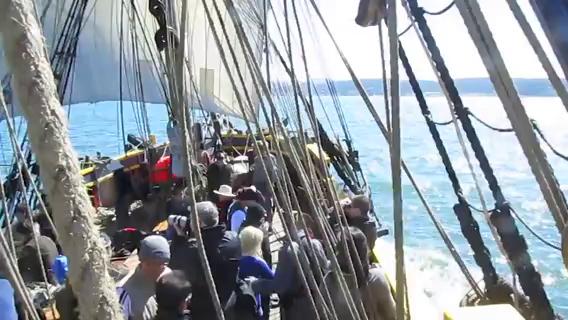} \hfill
        \includegraphics[height=0.15\linewidth]{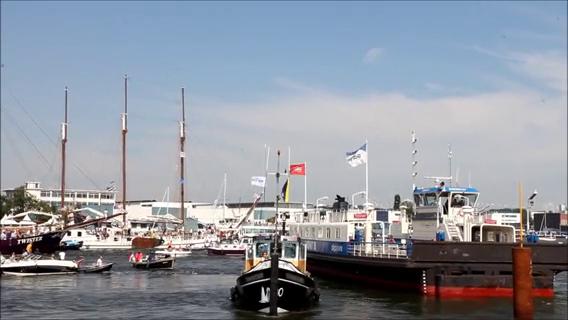}
        \\
        \includegraphics[height=0.15\linewidth]{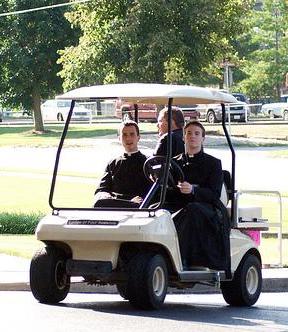} &
        \includegraphics[height=0.15\linewidth]{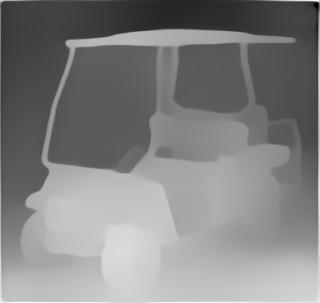} \hfill
        \includegraphics[height=0.15\linewidth]{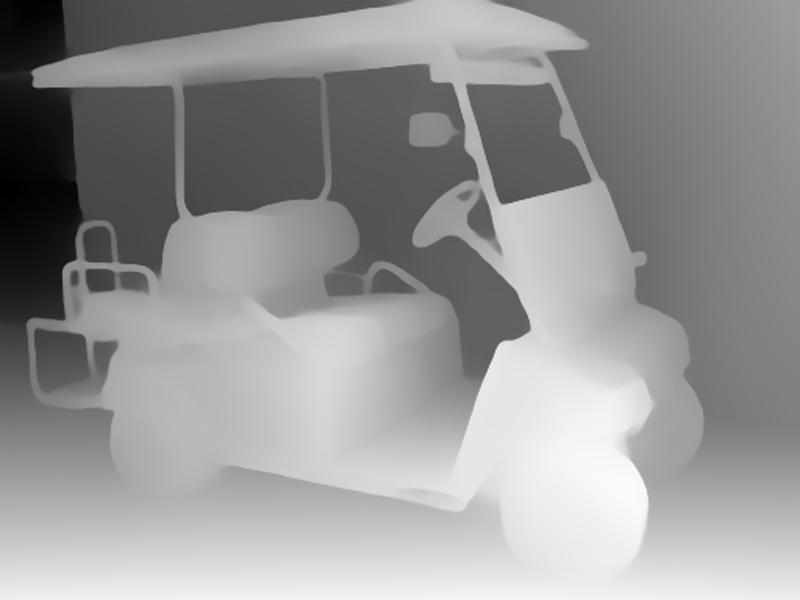} \hfill
        \includegraphics[height=0.15\linewidth]{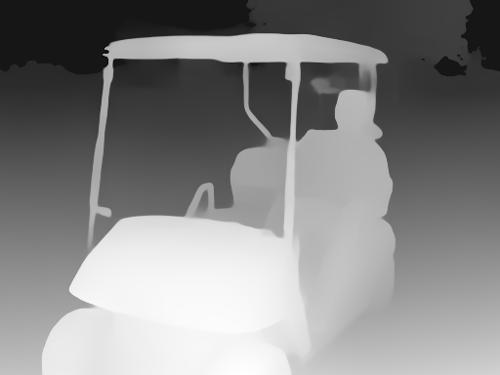} &
        \includegraphics[height=0.15\linewidth]{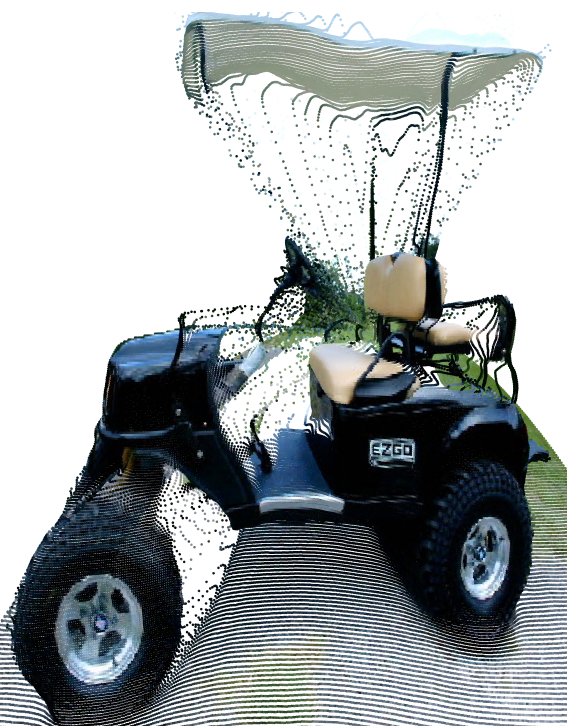} \hfill
        \includegraphics[height=0.15\linewidth]{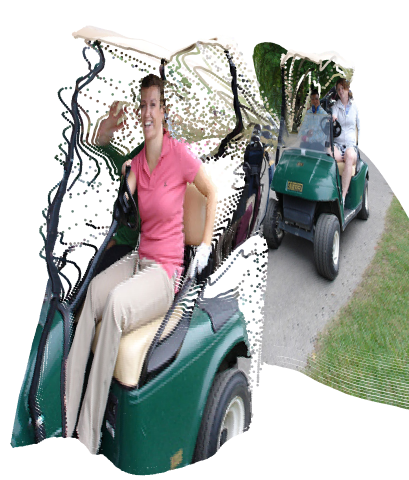} \hfill
        \includegraphics[height=0.15\linewidth]{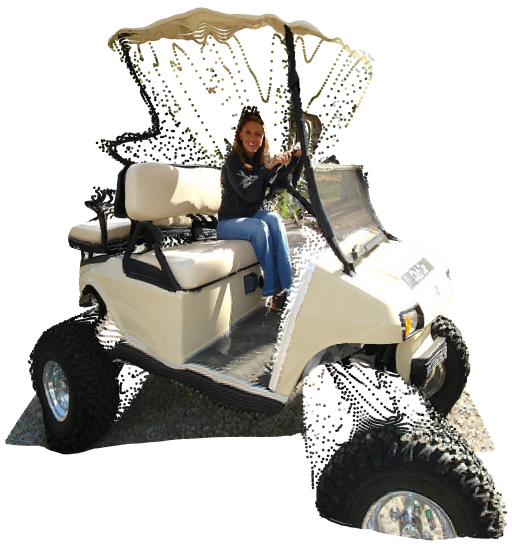} &
        \includegraphics[height=0.15\linewidth]{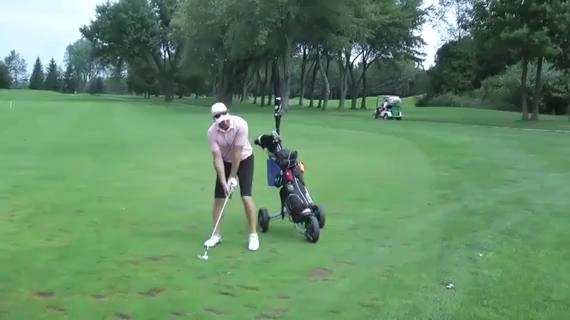} \hfill
        \includegraphics[height=0.15\linewidth]{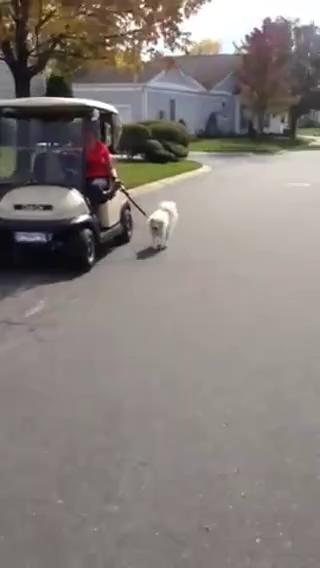} \hfill
        \includegraphics[height=0.15\linewidth]{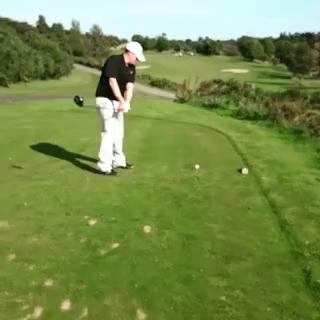}
    \end{tabular}%
}

    \vspace{-0.1in}
    \captionof{figure}{{\bf \OURS is a single vision model for many different visual modalities.} It learns to construct representations that are aligned across visual modalities, without requiring training data that specifies correspondences between those modalities. Using \OURS's shared visual representation, we successfully identify nearest neighbors of \textbf{left:} an image {\scriptsize (\imnet validation set)} in vision datasets that contain \textbf{right:} depth maps {\scriptsize (\imnet training set)}, single-view 3D images {\scriptsize (\imnet training set)}, and videos {\scriptsize (\kinetics validation set)}.
    }
    \label{fig:teaser}
\end{center}%
}]

\makeatletter{\renewcommand*{\@makefnmark}{}
\footnotetext{$^*$Equal technical contribution.}\makeatother}

\begin{abstract}
\vspace{-0.1in}
Prior work has studied different visual modalities in isolation and developed separate architectures for recognition of images, videos, and 3D data.
Instead, in this paper, we propose a single model which excels at classifying images, videos, and single-view 3D data using exactly the same model parameters.
Our `\OURS' model leverages the flexibility of transformer-based architectures and is trained jointly on classification tasks from different modalities.
\OURS is simple to train, uses off-the-shelf standard datasets, and performs at-par or better than modality-specific models of the same size.
A single \OURS model obtains 86.0\% on ImageNet, 84.1\% on Kinetics, and 67.1\% on SUN RGB-D.
After finetuning, our models outperform prior work on a variety of vision tasks and generalize across modalities.
\OURS's shared visual representation naturally enables cross-modal recognition without access to correspondences between modalities.
We hope our results motivate researchers to model visual modalities together.
\end{abstract}
\section{Introduction}
\label{sec:intro}

Computer vision research spans multiple modalities related to our perception of the visual world, such as images, videos, and depth.
In general, we study each of these modalities in isolation, and tailor our computer vision models to learn the best features from their specificities.
While these modality-specific models achieve impressive performance, sometimes even surpassing humans on their specific tasks, they do not possess the flexibility that a human-like vision system does---the ability to work across modalities.
We argue that the first step towards a truly all-purpose vision system is to build models that work seamlessly across modalities, instead of being over-optimized for each modality. %

Beyond their flexibility, such modality-agnostic models
have several advantages over their traditional, modality-specific counterparts.
First, a modality-agnostic model can perform \emph{cross-modal generalization}: it can use what it has learned from one modality to perform recognition in other modalities.
For example, it can recognize pumpkins in 3D images even if it has only seen labeled videos of pumpkins.
In turn, this allows existing labeled datasets to be used more effectively: it becomes possible to train models on the union of vision datasets with different input modalities.
Second, it {\em saves the research and engineering effort} spent on optimizing models for a specific modality.
For example, image and video models have followed a similar trajectory of evolution, from hand-crafted descriptors~\cite{lowe2004distinctive,laptev2003space} to convolutional networks~\cite{he2016deep,tran2018closer} and, eventually, vision transformers~\cite{dosovitskiy2020image,arnab2021vivit}; however, each had to be developed and tuned individually.
A common architecture would make scientific progress readily available to users of any visual modality.
Finally, a model that operates on many visual modalities is naturally multi-modal and can easily {\em leverage new visual sensors} as they becomes available.
For instance, a modality-agnostic recognition model running on a robot can readily exploit a new depth sensor when it is installed on that robot.
Despite such clear advantages, modality-agnostic models have rarely been studied and their performance compared to their modality-specific counterparts has been disappointing.
There are many reasons that explain this situation, such as the need for a flexible architecture with enough capacity to learn modality-specific cues from the different modalities; and enough compute to train it on video, images, and single-view 3D simultaneously.

This paper develops a modality-agnostic vision model that leverages recent advances in vision architectures~\cite{dosovitskiy2020image,liu2021swin}.
The model we develop is ``omnivorous'' in that it works on three different visual modalities: images, videos, and single-view 3D.
Our \OURS model does not use a custom architecture for each visual modality.
It performs recognition on all three modalities using the same, shared model parameters.
It works by converting each input modality into embeddings of spatio-temporal patches, which are processed by exactly the same Transformer~\cite{vaswani2017attention} to produce a representation of the input.
We train \OURS on a collection of standard, off-the-shelf classification datasets that have different input modalities.
Unlike prior work~\cite{gupta2013perceptual,simonyan2014two}, our training does not use explicit correspondences between different input modalities.

Our experiments demonstrate the advantages of our \OURS models.
Surprisingly, we find that \OURS representations generalize well across visual modalities (see~\cref{fig:teaser}) even though \OURS was not explicitly trained to model cross-modal correspondences.
These capabilities emerge without explicit cross-modal supervision simply due to the parameter sharing between models for different modalities.
On standard image, video, and single-view 3D benchmarks, \OURS performs at par with or better than modality-specific vision models with the same number of parameters.
The same \OURS model obtains 85.6\% top-1 accuracy on \imnet, 83.4\% top-1 on \kinetics, and 67.4\% top-1 accuracy on \sunrgbd.
\OURS's strong generalization capabilities also extend to transfer learning experiments. %
\OURS performs at par with recent large transformers on \imnet, sets a new state-of-the-art on action recognition benchmarks such as \epic, \sthsth, and on single-view 3D classification and segmentation benchmarks.
We believe our work presents a compelling argument for shifting towards the development of vision models that can operate on any visual modality.

\begin{figure}[!t]
    \centering
    \includegraphics[width=\linewidth]{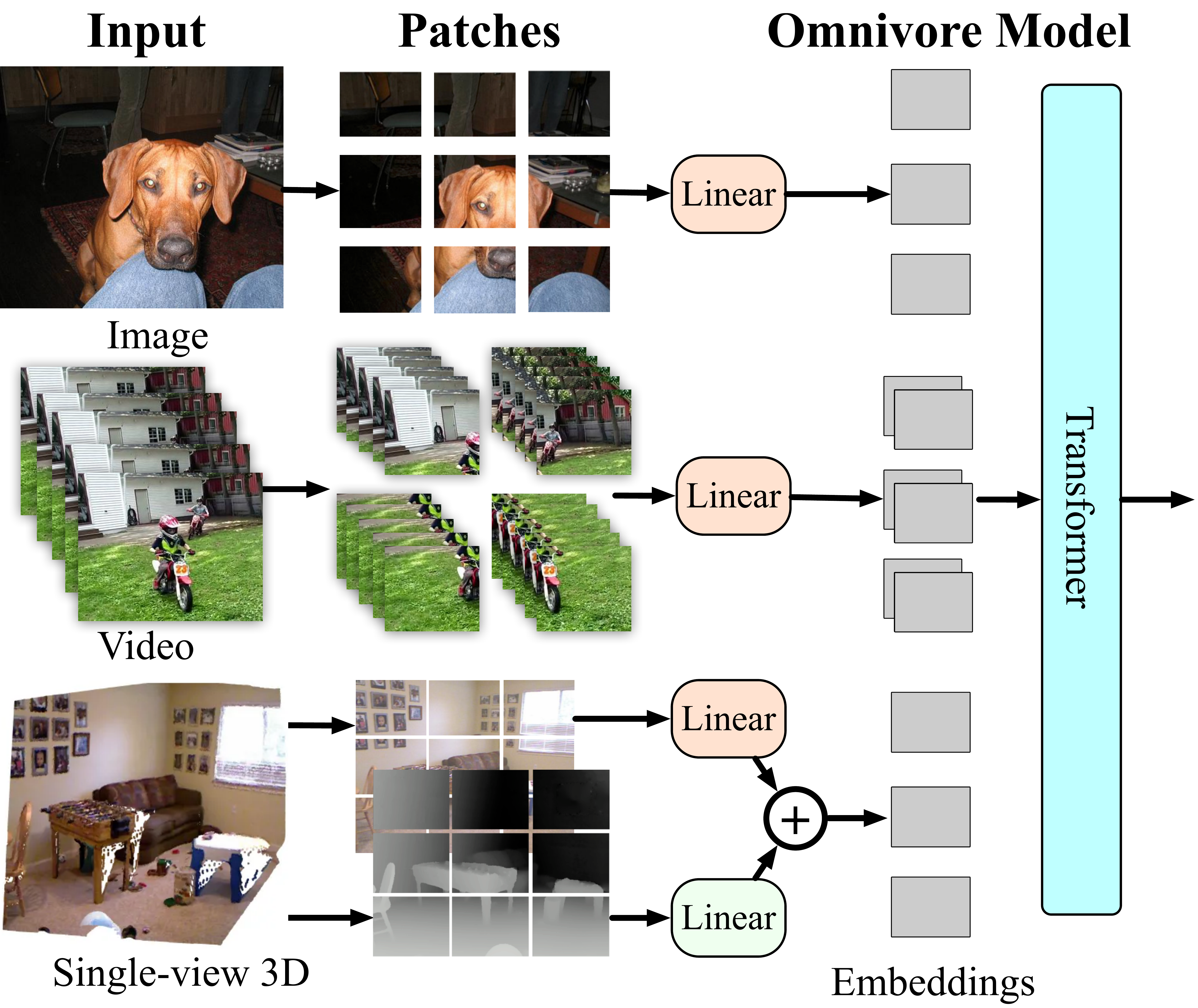}
    \vspace{-0.1in}
    \caption{
    \textbf{Multiple visual modalities in the \OURS model.}
    We convert image, video, and single-view 3D modalities into embeddings that are fed into a Transformer model.
    The images are converted into patches, videos into spatio-temporal tubes, and the single-view 3D images are converted into RGB patches and depth patches.
    The patches are projected into embeddings using linear layers. We use the same linear layer for (image or video) RGB patches and a separate one for depth patches.
    }
    \label{fig:approach}
\end{figure}

\section{Related Work}
\label{sec:related_work}
We build on prior work in ConvNet architectures, Transformers, multi-modal learning, and multi-task learning.

\par \noindent \textbf{ConvNet architectures in vision.}
ConvNet architectures~\cite{lecun1998gradient,fukushima1980self} have been popular for many computer vision tasks in images, video, and 3D recognition.
2D convolutions are the main building block in ConvNets for images~\cite{krizhevsky2012imagenet,szegedy2015going,simonyan2014two,he2016deep},
whereas 3D convolutions are used on 3D data~\cite{graham20183d,choy20194d} or are combined with 2D convolutions for recognition of videos~\cite{tran2015learning,tran2018closer,carreira2017quo}.
I3D~\cite{carreira2017quo} introduced a way to ``inflate'' 2D image convolutions into 3D convolutions, which allows 3D ConvNets for videos and 3D data to leverage image data indirectly via initialization from pretrained image models.
Since video and 3D datasets are relatively small, they benefit from inflated pretrained image networks.
However, while the inflation technique is applicable only to model finetuning, \OURS models are pretrained jointly on images, videos, and single-view 3D data.

\par \noindent \textbf{Transformers in vision.}
The Transformer architecture~\cite{vaswani2017attention} originally proposed for NLP tasks has been successfully applied in
computer vision on images~\cite{wang2021pyramid,wang2018non,carion2020end,dosovitskiy2020image,touvron2021training,parmar2018image},
video~\cite{arnab2021vivit,bertasius2021is,liu2021video,girdhar2019video,neimark2021video,girdhar2021anticipative},
and 3D data~\cite{pan20213d,zhao2020point,misra2021-3detr}.
Models such as ViT~\cite{dosovitskiy2020image}, Swin~\cite{liu2021swin}, and MViT~\cite{fan2021multiscale} perform competitively on benchmark tasks such as image classification, detection, and video recognition.
For example, Swin~\cite{liu2021swin,liu2021video} and MViT~\cite{fan2021multiscale} require minimal changes to be used in image or video recognition tasks.
Similarly, the Perceiver~\cite{jaegle2021perceiver} can model image, point cloud, audio, and video inputs.
However, all these studies train separate models for each visual modality.
Instead, we train a single model on multiple input modalities simultaneously, which equips our model with cross-modal generalization capabilities.

\par \noindent \textbf{Multi-modal learning.}
Our work uses multiple visual modalities to train the model.
Multi-modal learning architectures may involve training separate encoders for each type of input modality.
For example, a range of tasks require training separate encoders for images and text~\cite{miech2020end,gong2014improving,karpathy2015deep,lu202012,castrejon2016learning},
for video and audio~\cite{morgado2021audio,morgado2021robust,arandjelovic2017look,arandjelovic2018objects,owens,patrick2020multi}, or for video and optical flow~\cite{simonyan2014two}.
Recently, Transformers have been used to fuse multiple modalities:
Transformers have been used to fuse features in vision-and-language tasks~\cite{kamath2021mdetr,li2020oscar,hu2021unit,chen2020uniter,lu2019vilbert,tan2019lxmert,su2019vl,alayrac2020self}
and video-and-audio tasks~\cite{nagrani2021attention}, video-and-image tasks~\cite{bain2021frozen}, and even tasks that involve video, audio, and text~\cite{akbari2021vatt}.
Unlike our work, most prior work assumes that all input modalities are in correspondence and available simultaneously, which restricts them to using only multi-modal datasets.
In our work, we train a single model on different visual modalities without assuming simultaneous access to all modalities.
This allows us to leverage standard off-the-shelf single-modality vision datasets and we show that using a single shared encoder naturally leads to cross-modal generalization.

\par \noindent \textbf{Multi-task learning.}
Our work is also related to studies on multi-task learning~\cite{caruana1997}, which develop models that output predictions for multiple tasks on the same input~\cite{eigen2015predicting,kokkinos2017ubernet,maninis2019attentive,zhang2014facial,misra2016cross,ghiasi2021multi}.
Such multi-task learners are known to work well when the target tasks exhibit strong similarities~\cite{ZamirSSGMS18,misra2016cross}.
They differ from \OURS in that they operate on a single input modality but are trained to perform multiple tasks.
By contrast, our models are trained to perform a single task (\emph{i.e.}, classification) on a variety of input modalities.
Other multi-task learners operate on multi-modal inputs~\cite{kaiser2017one}, but they use hand-designed model components for each modality.

\section{Approach}
\label{sec:approach}

Our goal is to learn a single model that can operate on three major visual modalities: images, videos, and single-view 3D.
Because the model's input modalities have different sizes and layouts---videos have a temporal axis and single-view 3D has an extra depth channel---this poses a challenge in designing the model.
To overcome this challenge, we adopt the Transformer~\cite{vaswani2017attention} architecture because the self-attention mechanism gracefully handles variable-sized inputs.
\cref{fig:approach} presents an overview of our approach.

\subsection{The \textbf{\OURS} Model}
We convert all visual modalities into a common format by representing them via embeddings.
Our model then uses a series of spatio-temporal attention operations to construct a unified representation of the different visual modalities.

\par \noindent \textbf{Input patches.} We represent the different types of visual input as a 4D tensor $\bX \in \mathbb{R}^{T\times H \times W \times C}$,
where $T$ is the size of the temporal dimension, $H$ and $W$ of the spatial dimensions, and $C$ of the channel dimension.
Thus, RGB images $\bI \in \mathbb{R}^{1\times H\times W\times 3}$ have $T\!=\!1$ frame with $C\!=\!3$ channels,
RGB videos $\bV \in \mathbb{R}^{T\times H \times W \times 3}$ have $T\!>\!1$ frames, and
single-view 3D images $\bD \in \mathbb{R}^{1\times H\times W\times 4}$ have $T\!=\!1$ frame with three RGB channels and one depth channel.

We follow~\cite{liu2021swin,dosovitskiy2020image,liu2021video} and split the input into a collection of patches.
We illustrate this process in \cref{fig:approach}.
Specifically, we convert the visual input $\bX$ into a set of 4D sub-tensors $\bx$ of size $t\times {h}\times {w}\times c$.
Images $\bI$ are split into a set of non-overlapping image patches of size $1\times {h}\times {w}\times 3$.
Similarly, videos $\bV$ are split into a set of non-overlapping spatio-temporal patches of shape $t\times {h}\times {w}\times 3$.
For single-view 3D images $\bD$, the image (RGB) and depth (D) channels are converted separately into patches of size $1\times {h}\times {w}\times 3$ and $1\times {h}\times {w}\times 1$, respectively.

\par \noindent \textbf{Model architecture.}
\noindent Our model $f$ maps the resulting spatio-temporal visual patches
into a shared representation $\mathbf{\Phi}$ for images, videos, and single-view 3D.
We design the model to enable maximal parameter sharing across visual modalities.
The input layer of the model processes each patch $\bx$ independently, and projects the patches into an embedding $\be$ using a linear layer followed by a LayerNorm~\cite{ba2016layer} (linear+LN). %
Each patch $\bx$ of shape $t\times {h}\times {w}\times c$ is converted into an embedding of size $d$.
We use the same layers to embed all the three-channel RGB patches, \ie, for image patches, video patches, and patches of the first three channels of a single-view 3D image.
We zero-pad the single-frame patches on one side to ensure all patches have the same shape, $t\times {h}\times {w}\times 3$.
We use a separate linear+LN layer to embed the depth-channel patches and add its output to the embedding of the corresponding RGB patch.

We use the same model (parameters) to process all the resulting embeddings.
While \OURS can use any vision transformer architecture~\cite{fan2021multiscale,dosovitskiy2020image} to process the patch embeddings, we use the Swin transformer architecture~\cite{liu2021swin} as our base model given its strong performance on image and video tasks.
We rely on the self-attention~\cite{vaswani2017attention} operation for spatio-temporal modeling across the patch embeddings, $\be$.
Akin to~\cite{liu2021swin}, the self-attention involves patch embeddings from spatially and temporally nearby patches.
We also use two sets of relative positional encodings: one for the spatial dimension and the other for the temporal dimension.

\subsection{Training the \textbf{\OURS} Model}
The \OURS model $f$ creates a single embedding $f(\bX) = \bPhi$ for multiple types of visual inputs.
We train our model using a collection of classification tasks that provide inputs $\{(\bX_{i}, y_{i})\}$ with a visual input, $\bX_{i}$, and a label, $y_{i}$.
For example, we train most \OURS models jointly on the \imnet dataset for image classification, the \kinetics dataset for action recognition, and the \sunrgbd dataset for single-view 3D scene classification.

This approach is similar to multi-task learning~\cite{caruana1997} and cross-modal alignment~\cite{castrejon2016learning}, but there important differences.
In particular, we neither assume that the input observations are aligned (\ie, we do not assume access to correspondences between images, videos, and 3D data) nor do we assume
that these datasets share the same label space.
To achieve this, we employ dataset-specific linear classification layers on top of the final representation, $\bPhi$, produced by the model.
The training loss of a sample is computed based solely on the output of the classification layer that corresponds to that sample's source dataset.

\par \noindent \textbf{Loss and optimization.}
We train \OURS to minimize the cross-entropy loss on the training datasets using mini-batch SGD.
We experiment with two different mini-batch construction strategies for SGD.
In our first strategy, we construct mini-batches from each dataset (modality) separately.
This strategy is easy to implement but alternating between datasets may potentially lead to training instabilities.
Hence, we experiment with a second strategy
that
constructs mini-batches that mix samples from all datasets.
We evaluate both mini-batch construction strategies in~\cref{sec:ablations}.

\section{Experiments}
\label{sec:experiments}

We perform a series of experiments to assess the effectiveness of \OURS.
Specifically, we compare \OURS models to their modality-specific counterparts and to \sota models on a variety of recognition tasks.
We also ablate several design choices we made in \OURS.

\par \noindent \textbf{Pre-training datasets.}
We train \OURS on images from the \imnet dataset~\cite{ILSVRC15}, videos from the Kinetics dataset~\cite{kay2017kinetics}, and single-view 3D images from the \sunrgbd dataset~\cite{song2015sun}.
We measure the top-1 and top-5 classification accuracy of our models on the respective validation sets.
We note that the three datasets have negligible overlap in their visual concepts: \imnet focuses on object-centric classes, \kinetics on action classes, and \sunrgbd on indoor scene classes.

\par \noindent \underline{\emph{Images.}} The \imnet (\imnetShort) dataset has \app1.2M training and 50K validation images that comprise 1,000 classes.

\par \noindent \underline{\emph{Videos.}}
The \kinetics (\kineticsShort) dataset consists of \app240K training and 20K validation video clips that are 10 seconds long, and are labeled into one of 400 action classes.
\par \noindent \underline{\emph{Single-view 3D.}} The \sunrgbd dataset has \app5K train and \app5K val RGBD images with 19 scene classes.
Following~\cite{ranftl2020towards}, we convert the depth maps into disparity maps.

\par \noindent \textbf{Implementation details.}
We use the Swin transformer~\cite{liu2021swin,liu2021video} architecture as the backbone for \OURS, and attach linear heads for each target dataset.
At training time, we use a resolution of $224\!\times\!224$ and train using standard image augmentations~\cite{touvron2021training} on ImageNet.
For Kinetics, we sample 32 frames at stride 2. %
\sunrgbd is processed similarly to ImageNet but we randomly drop the RGB channels with a probability of $0.5$ in order to encourage the model to use the depth channel for recognition as well.
We provide complete implementation details in~\cref{app:pretraining_details}.
Our models are optimized using AdamW~\cite{loshchilov2017decoupled} for 500 epochs where a single epoch consists of one epoch each for \imnet and Kinetics, and 10 epochs for \sunrgbd.

\par \noindent \textbf{Transfer datasets and metrics.}
We evaluate \OURS in transfer learning experiments on a diverse set of image, video, and single-view 3D tasks; see~\cref{tab:transfer_datasets} for a summary.
We present details on the experimental setup in~\cref{app:details_transfer}.

\begin{table}
    \centering
    \resizebox{\linewidth}{!}{
    \setlength{\tabcolsep}{2pt}
    \begin{tabular}{l|cccc}
        \bf Dataset  & \bf Task & \bf \#cls & \bf \#train & \bf \#val \\
        \hline
        {\color{ImageDark}{\inat}} (\inatShort)~\cite{iNaturalist} & Fine-grained cls. & 8142 & 437K & 24K \\
        {\color{ImageDark}{\pets}} (\petsShort)~\cite{parkhi2012cats} & Fine-grained cls. & 37 & 3.6K & 3.6K \\
        {\color{ImageDark}{\placesThree}} (\placesThreeShort)~\cite{zhou2017places} & Scene cls. & 365 & 1.8M & 36K \\
        {\color{VideoDark}{\sthsth}} (\sthsthShort)~\cite{goyal2017something} & Action cls. & 174 & 169K & 25K \\
        {\color{VideoDark}{\epic}} (\epicShort)~\cite{damen2021rescaling} & Action cls. & 3806 & 67K & 10K \\
        {\color{ThreeDDark}{\nyu}} (\nyuShort)~\cite{Silberman_ECCV12_NYUv2} & Scene cls. & 10 & 794 & 653 \\
        {\color{ThreeDDark}{\nyuSeg}} (\nyuSegShort)~\cite{Silberman_ECCV12_NYUv2} & Segmentation & 40 & 794 & 653 \\
    \end{tabular}
    }
    \caption{\textbf{Transfer datasets} used to evaluate \OURS on {\color{ImageDark} image}, {\color{VideoDark} video} and {\color{ThreeDDark} single-view 3D} modalities.
    The table reports the task, number of classes (\#cls), number of training samples (\#train), and number of validation samples (\#val) for each dataset.
    }
    \label{tab:transfer_datasets}
\end{table}

\par \noindent \underline{\emph{Images.}}
We evaluate \OURS on fine-grained object recognition on the \inat dataset~\cite{iNaturalist}, fine-grained classification on the \pets dataset~\cite{parkhi2012cats}, and in scene classification on the \placesThree dataset~\cite{zhou2017places}.

\par \noindent \underline{\emph{Videos.}}
We use the \sthsth dataset, which has a special emphasis on temporal modeling for action recognition.
We also use the \epic dataset, which has 100 hours of unscripted egocentric video.
Each clip is labeled with a verb and a noun that together form an action.
Our model is trained to recognize all 3,806 actions, \ie, verb-noun pairs in the dataset. We marginalize over verbs to obtain noun predictions and vice versa.

\par \noindent \underline{\emph{Single-view 3D.}}
We use the \nyu dataset for single-view 3D scene classification and segmentation.
We follow the setup from~\cite{gupta2013perceptual} for scene classification and~\cite{gupta2013perceptual,cao2021shapeconv} for segmentation.
For segmentation, we follow~\cite{liu2021swin} and use the UPerNet~\cite{xiao2018unified} head with the Swin trunk.

\subsection{Comparison with Modality-Specific Models}
\label{sec:in1k_comparisons}

\begin{table}[t]
    \centering
    \resizebox{\columnwidth}{!}{
    \begin{tabular}{l | cc | cc | cH}
    \rowcolor{white}& \multicolumn{2}{i}{ \color{ImageDark} \bf\imnet} & \multicolumn{2}{v}{ \color{VideoDark} \bf\kinetics} & \multicolumn{1}{t}{ \color{ThreeDDark} \bf\sunrgbdShort} \\
    \rowcolor{white}\multirow{-2}{*}{\bf Method}& {\color{ImageDark}top-1} & {\color{ImageDark}top-5} & {\color{VideoDark}top-1} & {\color{VideoDark}top-5} & {\color{ThreeDDark}top-1}\\ %
    \midrule
    Image\swinT~\cite{liu2021swin} & \bf 81.2 & 95.5 &  \xmark &  \xmark &  \xmark &  \xmark \\
    Video\swinT~\cite{liu2021video} &  \xmark &  \xmark & 78.8 & 93.6 &  \xmark &  \xmark \\
    Depth\swinT & \xmark &  \xmark &  \xmark  &  \xmark & \bf 63.1 & 90.7 \\ %
    \OURS (\swinT) & 80.9 & \bf 95.5 & \bf 78.9 & \bf 93.8 & 62.3 & 87.5 \\  %
    \hline

    Image\swinS~\cite{liu2021swin} & 83.2 & 96.2 &  \xmark &  \xmark &  \xmark &  \xmark \\
    Video\swinS~\cite{liu2021video} &  \xmark &  \xmark & 80.6 & 94.5 &  \xmark &  \xmark \\  %
    Depth\swinS & \xmark &  \xmark &  \xmark  &  \xmark & \bf 64.9 & 91.6 \\ %
    \OURS (\swinS) & \bf 83.4 & \bf 96.6 & \bf 82.2 & \bf 95.4 & 64.6 & 89.1 \\  %
    \hline

    Image\swinB~\cite{liu2021swin} & 83.5 & 96.5 &  \xmark &  \xmark &  \xmark &  \xmark \\
    Video\swinB~\cite{liu2021video} &  \xmark &  \xmark & 80.6 & 94.6 &  \xmark &  \xmark \\
    Depth\swinB &  \xmark &  \xmark &  \xmark &  \xmark & 64.8 & 92.5 \\ %
    \OURS (\swinB) & \bf 84.0 & \bf 96.8 & \bf 83.3 & \bf 95.8 & \bf 65.4 & 89.8 \\ %

    \arrayrulecolor{black}
    \end{tabular}%
    }
    \caption{{\bf \OURS \vs modality-specific models} that have the same model architecture and number of parameters.
    \OURS is a single model trained from scratch jointly on the \imnetShort, \kineticsShort and \sunrgbdShort datasets whereas
    the modality-specific models are trained specifically for each dataset (modality).
    The ImageSwin model is trained from scratch while the VideoSwin and DepthSwin models are finetuned from the ImageSwin model.
    \OURS performs at-par or outperforms modality-specific models.
    }
    \label{tab:inflation_vs_us}
\end{table}

\begin{figure*}
    \begin{minipage}[t]{0.285\linewidth}
        \vspace{0pt}
        \centering
        \includegraphics[width=\linewidth]{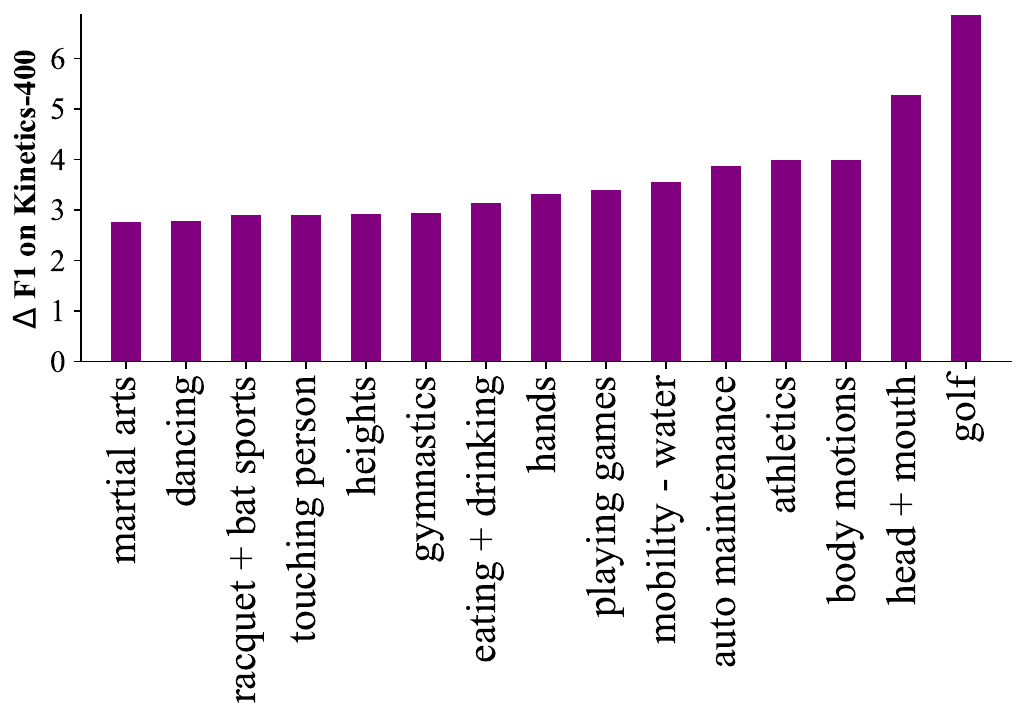}%
        \vspace{-0.1in}
        \caption{\textbf{Comparing \OURS with VideoSwin} on \kineticsShort.
        \OURS improves over VideoSwin on F1 score on all 38 class groups defined in~\cite{kay2017kinetics} (top 15 shown here for brevity).
        }\label{fig:per_group_gain_k400}
    \end{minipage}\hfill
    \begin{minipage}[t]{0.70\linewidth}
        \vspace{0pt}
        \setlength{\tabcolsep}{3pt}
        \centering
        \resizebox{\linewidth}{!}{
        \begin{tabular}{l l | cc cc cc | cc cc | c c}
        \rowcolor{white} & & \multicolumn{2}{i}{ \color{ImageDark} \bf \placesThreeShort} & \multicolumn{2}{i}{ \color{ImageDark} \bf \inatShort} &  \multicolumn{2}{i}{ \color{ImageDark} \bf \petsShort} &
        \multicolumn{2}{v}{ \color{VideoDark} \bf \sthsthShort} & \multicolumn{2}{v}{ \color{VideoDark} \bf \epicShort}
        & \multicolumn{1}{t}{ \color{ThreeDDark} \bf\nyuShort} & \multicolumn{1}{t}{ \color{ThreeDDark} \bf\nyuSegShort} \\
    \rowcolor{white}\multirow{-2}{*}{\bf Model} & \multirow{-2}{*}{\bf Method} & {\color{ImageDark} top-1} & {\color{ImageDark} top-5}  & {\color{ImageDark} top-1} & {\color{ImageDark} top-5} & {\color{ImageDark} top-1} & {\color{ImageDark} top-5} &
        {\color{VideoDark}  top-1} & {\color{VideoDark}  top-5} & {\color{VideoDark}  top-1} & {\color{VideoDark}  top-5} & {\color{ThreeDDark} top-1} & %
        { \color{ThreeDDark} mIoU}  \\
        \midrule
        \multirow{2}{*}{\bf\swinT} & Specific &
        57.9 & 87.3 & %
        \bf 69.7 & 87.6 &
        93.7 & 99.6 & %
        62.2 & 88.7 &  %
        41.8 & 62.8 &  %
        72.5 & %
        47.9 \\ %

        & \OURS &
        \bf 58.2 & \bf 87.4 & %
        69.0 & \bf 87.7 & %
        \bf 94.2 & \bf 99.7 & %
        \bf 64.4 & \bf 89.7 &  %
        \bf 42.7 & \bf 63.1 &  %
        \bf 77.3 & %
        \bf 49.7 \\
        \hline

        \multirow{2}{*}{\bf\swinS} & Specific &
        58.7 & \bf 88.1 & %
        72.9 & 90.2 & %
        94.4 & 99.6 & %
        66.8 & 91.1 & %
        42.5 & 63.4 &  %
        76.7 & %
        51.3 \\ %

        & \OURS &
        \bf 58.8 & 88.0 & %
        \bf 73.6 & \bf 90.8 & %
        \bf 95.2 & \bf 99.7 & %
        \bf 68.2 & \bf 91.8 &  %
        \bf 44.9 & \bf 64.8 &  %
        \bf 76.9 & %
        \bf 52.7 \\
        \hline

        \hline
        \multirow{2}{*}{\bf\swinB} & Specific &
        58.9 & \bf 88.3 & %
        73.2 & 90.9 & %
        94.2 & 99.7 & %
        65.8 & 90.6 & %
        42.8 & 64.0 &   %
        76.4 & %
        51.1 \\ %

        & \OURS &
        \bf 59.2 & \bf 88.3 & %
        \bf 74.4 & \bf 91.1 & %
        \bf 95.1 & \bf 99.8 & %
        \bf 68.3 & \bf 92.1 &  %
        \bf 47.4 & \bf 67.7 &  %
        \bf 79.4 & %
        \bf 54.0 \\
        \hline
        \arrayrulecolor{black}
        \end{tabular}}
        \captionof{table}{{\bf Comparing \OURS with modality-specific models} after finetuning the models on seven downstream tasks.
        Results are presented for three different model sizes: T, S, and B.
        Our {\color{ImageDark} image} specific model is pretrained on \imnetShort.
        The {\color{VideoDark} video} specific and {\color{ThreeDDark} single-view 3D} specific models are both initialized using inflation from the pretrained image-specific model and finetuned on \kineticsShort and \sunrgbd respectively.
        \OURS models are at par with or outperform modality-specific models on nearly all downstream tasks.
        }
        \label{tab:inflation_vs_us_finetuned}
    \end{minipage}
\end{figure*}

We compare \OURS to models trained on a specific visual modality.
We train \OURS from scratch jointly on the \imnetShort, \kineticsShort, and \sunrgbdShort datasets.
Our modality-specific baseline models use the same Swin transformer architecture as \OURS; we refer to them as ImageSwin, VideoSwin, and DepthSwin.
Excluding the patch-embedding linear layers, these models have the same number of parameters as \OURS.
Following standard practice~\cite{liu2021video,liu2021swin}, the ImageSwin model is trained on \imnetShort, whereas VideoSwin and DepthSwin models are finetuned by inflating the ImageSwin model.
We experiment with three model sizes: \emph{viz.}, \swinT, \swinS, and \swinB.\footnote{We refer to~\cite{liu2021swin} for details on these model sizes.}

\par \noindent \textbf{Pretraining performance.}
In~\cref{tab:inflation_vs_us}, we compare \OURS to modality-specific models on the pretraining datasets.
The results in the table show that across model sizes, \OURS models match or exceed the performance of their modality-specific counterparts.
This observation supports our hypothesis that it is possible to learn a single visual representation that works across visual modalities.
\OURS learns representations that are as good as modality-specific representations using the same training data, same model parameters and same model capacity.
This implies that \OURS provides a viable alternative to the pretrain-then-finetune paradigm commonly used to deploy modality-specific models: it can deliver the same or better recognition accuracy with a third of the parameters. %

From our results, we also observe that higher-capacity models benefit more from omnivorous training.
\OURS models using the larger \swinB architecture improve over their modality-specific counterparts on both \imnetShort and \kineticsShort, whereas the smallest \swinT model does not.

\cref{fig:per_group_gain_k400} presents a detailed analysis of the improvements of \OURS over the VideoSwin baseline (both using the \swinB architecture) on the \kineticsShort dataset.
Here VideoSwin is pre-trained on \imnetShort and finetuned on \kineticsShort, whereas \OURS is trained jointly on \imnetShort, \kineticsShort, and \sunrgbd. Both models use the the \swinB architecture.
\OURS particularly improves the recognition of classes that require reasoning about parts of the human body such as the hands, arms, head, mouth, hair \etc.
We surmise this is because joint training on images helps \OURS to learn a better model of the spatial configuration of parts.

\par \noindent \textbf{Transfer learning performance.}
We compare \OURS to modality-specific models by finetuning on various downstream tasks.
\cref{tab:inflation_vs_us_finetuned} presents the results of these experiments.
We observe that \OURS transfers better than modality-specific models on nearly all downstream tasks.
In particular, \OURS provides significant gains on video-recognition tasks, even though it does not get any additional video supervision during pre-training compared to the baseline.
We reiterate that \OURS %
has the same model capacity as the modality-specific baselines.
This observation underscores one of the key benefits of multi-modal training: because \OURS was pretrained jointly on more diverse training data, it generalizes better out-of-distribution.
As before, \cref{tab:inflation_vs_us_finetuned} also shows that higher-capacity models benefit the most from omnivorous training.

\begin{table}[t]
    \centering
    \resizebox{\columnwidth}{!}{
    \begin{tabular}{l | cc | cc | cH}
    \rowcolor{white}& \multicolumn{2}{i}{ \color{ImageDark} \bf\imnet} & \multicolumn{2}{v}{ \color{VideoDark} \bf\kinetics} & \multicolumn{1}{t}{ \color{ThreeDDark} \bf\sunrgbdShort} \\
    \rowcolor{white}\multirow{-2}{*}{\bf Method}& {\color{ImageDark}top-1} & {\color{ImageDark}top-5} & {\color{VideoDark}top-1} & {\color{VideoDark}top-5} & {\color{ThreeDDark}top-1} \\ %
    \midrule
    MViT-B-24~\cite{fan2021multiscale} & 83.1 & - & \xmark & \xmark & \xmark & \xmark \\
    ViT-L/16~\cite{dosovitskiy2020image} & 85.3 & - & \xmark & \xmark & \xmark & \xmark \\
    Image\swinB~\cite{liu2021swin} & 85.2 & 97.5 & \xmark & \xmark & \xmark & \xmark \\
    Image\swinL~\cite{liu2021swin} & 86.3 & 97.9 & \xmark & \xmark & \xmark & \xmark \\
    \hline
    ViT-B-VTN~\cite{neimark2021video} & \xmark & \xmark & 79.8  & 94.2 & \xmark & \xmark \\
    TimeSformer-L~\cite{bertasius2021is} & \xmark & \xmark & 80.7 & 94.7 & \xmark & \xmark \\
    ViViT-L/16x2 320~\cite{arnab2021vivit} & \xmark & \xmark & 81.3 & 94.7 & \xmark & \xmark \\
    MViT-B 64$\times$3 ~\cite{fan2021multiscale} & \xmark & \xmark & 81.2 & 95.1 & \xmark & \xmark \\
    VideoSwin-B~\cite{liu2021video} & \xmark & \xmark & 82.7 & 95.5 & \xmark & \xmark \\
    VideoSwin-L~\cite{liu2021video} & \xmark & \xmark & 83.1 & 95.9 & \xmark & \xmark \\
    \hline
    DF$^2$Net~\cite{li2018df2net} & \xmark & \xmark & \xmark & \xmark & 54.6 & - \\
    G-L-SOOR~\cite{song2020image} & \xmark & \xmark & \xmark & \xmark & 55.5 & - \\
    TRecgNet~\cite{du2019translate} & \xmark & \xmark & \xmark & \xmark & 56.7 & - \\
    CNN-RNN~\cite{caglayan2020CNNrandRNN} & \xmark & \xmark & \xmark & \xmark & 60.7 & 89.9 \\  %
    Depth \swinB & \xmark & \xmark & \xmark & \xmark & 69.1 & 93.2 \\ %
    Depth \swinL & \xmark & \xmark & \xmark & \xmark & 68.7 & 93.8 \\
    \hline

    \OURS (\swinB) & 85.3 & 97.5 & 84.0 & 96.2 & 67.2 & 92.0 \\ %

    \OURS (\swinL) & 86.0 & 97.7 & 84.1 & 96.3 & 67.1 & 91.4 \\ %
    \arrayrulecolor{black}
    \end{tabular}%
    }
    \caption{{\bf Comparing \OURS with \sota models} %
    on the {\color{ImageDark}image}, {\color{VideoDark}video}, and {\color{ThreeDDark}single-view 3D} classification datasets used to pre-train \OURS.
    \OURS performs on par with or better than \sota models on all three pre-training tasks, including modality-specific models of similar size.
    }
    \label{tab:sota}
\end{table}

\subsection{Comparison with the state-of-the-art}
\label{sec:sota}
Next, we perform experiments comparing \OURS to existing \sota models. %
In these experiments, like many \sota modality-specific methods, we use the \imnetFull (\imnetFullShort) dataset during pretraining.
The \OURS \swinB and \swinL models are trained from scratch on \imnetFullShort, \imnetShort, \kineticsShort, and \sunrgbdShort, where a single epoch consists of one epoch each of \imnetShort and \kineticsShort, 10 epochs of \sunrgbdShort, and 0.1 epochs of \imnetFull.
\cref{tab:sota} compares the performance of the \OURS models to \sota models on each of the three benchmarks.
\OURS performs at par with or exceeds modality-specific methods despite using a model architecture that is not tailored towards any specific modality.
Even when compared to modality-specific models with a similar number of parameters, \OURS models match the \sota on \imnetShort, and outperform the previous \sota on \kineticsShort by achieving 84.1\% accuracy -- a gain of 1\% which was previously only possible by using additional large video datasets.
This demonstrates the strong performance of using the same \OURS model across image, video and single-view 3D benchmarks.

\begin{table}
\centering
 \resizebox{0.75\columnwidth}{!}{
    \begin{tabular}[t]{l | ccc}
        \bf Method & {\color{ImageDark}\bf \placesThreeShort} &  {\color{ImageDark}\bf\inatShort} & {\color{ImageDark}\bf\petsShort} \\
        \midrule
        EfficientNet B6~\cite{xie2020adversarial, singh2022revisiting} & 58.5 & 79.1 & 95.4 \\ %
        EfficientNet B7~\cite{xie2020adversarial, singh2022revisiting} & 58.7 & 80.6 &  -- \\ %
        EfficientNet B8~\cite{xie2020adversarial, singh2022revisiting} & 58.6 & 81.3 &  -- \\ %
        DeiT-B~\cite{touvron2021training} $\uparrow$ & -- & 79.5 & -- \\
        ViT-B/16~\cite{dosovitskiy2020image, singh2022revisiting} $\uparrow$ & 58.2 & 79.8 & -- \\ %
        ViT-L/16~\cite{dosovitskiy2020image, singh2022revisiting} $\uparrow$ & 59.0 & 81.7 & -- \\ %
        \hline
        \OURS (\swinB) &  59.3 %
        & 76.3 &
        95.5
        \\
        \OURS (\swinB $\uparrow$) & 59.6
        & 82.6
        & 95.9
        \\
        \OURS (\swinL) & 59.4
        & 78.0
        & 95.7
         \\
        \OURS (\swinL $\uparrow$) & \bf 59.9 & \bf 84.1 &
        \bf 96.1
         \\
    \end{tabular}
 }
    \caption{\textbf{Comparing \OURS with \sota models in {\color{ImageDark} image} classification finetuning experiments} on three datasets.
    \OURS representations generalize well to scene classification (\placesThreeShort) and fine-grained classification (\inatShort, \petsShort).
    $\uparrow$ indicates finetuning on a higher resolution image ($384\!\times\!384$px; see~\cite{touvron2019fixing}).
    }
    \label{tab:sota-finetune-image}
\end{table}

\par \noindent \textbf{Transfer learning performance.}
We compare \OURS models to modality-specific models by finetuning on downstream tasks.
In~\cref{tab:sota-finetune-image}, we report results on image classification.
\OURS models outperform prior state-of-the-art in scene classification on \placesThree, and in fine-grained classification on \inat and \pets.

We finetune \OURS on video-classification and report the results in~\cref{tab:sota-finetune-video-cls}.
On the \epic dataset, the \OURS \swinB model %
achieves the absolute best performance across verb, noun, and verb-noun pair (action) classification.
Similarly, on the \sthsthShort dataset, which requires temporal reasoning, \OURS outperforms all prior work.
This suggests that \OURS representations transfer well to temporal-reasoning tasks -- \OURS sets a new \sota while outperforming architectures specialized for these video tasks.

Finally, in~\cref{tab:sota-finetune-nyu}, we report finetuning results for RGBD scene classification and segmentation.
While prior work relies on specialized 3D operators~\cite{cao2021shapeconv}, fusion techniques~\cite{yue2021two}, or depth encoding schemes~\cite{gupta2013perceptual}, \OURS uses a generic architecture and operates directly on disparity.
\OURS achieves \sota performance on both the scene classification and segmentation tasks.

\subsection{Ablation Study}
\label{sec:ablations}

\begin{table}
\centering
\resizebox{\columnwidth}{!}{
        \begin{tabular}[t]{l | cca|ac}
            & \multicolumn{3}{c}{\color{VideoDark}{\bf \epicShort}} & \multicolumn{2}{c}{\color{VideoDark}{\bf \sthsthShort}} \\
            \bf Method & verb & noun & action & top-1 & top-5 \\
            \hline
            \rowcolor{highlightRowColor} \multicolumn{6}{l}{\textit{RGB-only methods}} \\
            \hline
            SlowFast~\cite{feichtenhofer2019slowfast} & 65.6 & 50.0 & 38.5 & 63.0 & 88.5 \\
            TimeSformer~\cite{bertasius2021is} & -- & -- & -- & 62.4 & -- \\
            MViT-B-24~\cite{fan2021multiscale} & -- & -- & -- & 68.7 & 91.5 \\
            TAR~\cite{sener2021technical} & 66.0 & 53.4 & 45.3 & -- & -- \\
            VIMPAC~\cite{tan2021vimpac} & -- & -- & -- & 68.1 & -- \\
            ViViT-L~\cite{arnab2021vivit} & 66.4 & 56.8 & 44.0 & 65.9 & 89.9 \\
            MFormer-L~\cite{patrick2021keeping} & 67.1 & 57.6 & 44.1 & 68.1 & 91.2 \\
            ORViT~\cite{herzig2021object} & 68.4 & 58.7 & 45.7 & 69.5 & 91.5 \\
            {\sc CoVeR}~\cite{zhang2021cover} & -- & -- & -- & 70.9 & -- \\
            Video\swinB~\cite{liu2021video} & 67.8 & 57.0 & 46.1 & 69.6 & 92.7 \\  %
            {\bf \OURS} (\swinB) & \bf 69.5 & \bf 61.7 & \bf 49.9 & \bf 71.4 & \bf 93.5 \\ %
            \hline
            \rowcolor{highlightRowColor} \multicolumn{6}{l}{\textit{\demph{Multi-modal methods}}} \\
            \hline
            \demph{MML}~\cite{komkov2020mutual} & \demph{--} & \demph{--} & \demph{--} & \demph{69.1} & \demph{92.1} \\
            \demph{MTCN}~\cite{kazakos2021MTCN} & \demph{70.7} & \demph{62.1} & \demph{49.6} & \demph{--} & \demph{--} \\
        \end{tabular}}
\caption{\textbf{Comparing \OURS with \sota models in {\color{VideoDark} video} classification finetuning experiments} on two datasets.
We highlighted columns that show the two primary classification metrics used in prior work. \OURS models obtain \sota results on both datasets, even outperforming some multi-modal methods. %
}
\label{tab:sota-finetune-video-cls}
\end{table}

\begin{table}
    \centering
    \resizebox{\columnwidth}{!}{
    \begin{tabular}[b]{l | HHc | c}
        \bf Method & & & \bf Classification & \bf Segmentation \\
        \midrule
        DF$^2$Net~\cite{li2018df2net} & 61.1 & 54.8 & 65.4 & \xmark\\
        TRecgNet~\cite{du2019translate} & 64.8 & 57.7 & 69.2 & \xmark\\
        ShapeConv~\cite{cao2021shapeconv} & \xmark & \xmark & \xmark & 51.3 \\
        BCMFP + SA-Gate~\cite{chen2020-SAGate} & \xmark & \xmark & \xmark & 52.4 \\
        TCD~\cite{yue2021two} & \xmark & \xmark & \xmark & 53.1 \\
        \hline
        \OURS (\swinB) & 80.2 & 72.8 
        & 80.0 %
        & 55.1 \\
        \OURS (\swinL) & 79.8 & 72.7 
        & \bf 80.3 %
        & \bf 56.8 \\
    \end{tabular}
    }
    \caption{\textbf{Comparing \OURS with \sota models in {\color{ThreeDDark} RGBD} finetuning experiments} on the \nyu dataset.
    The left column shows the scene classification accuracy while the right column shows the mean intersection-over-union of semantic segmentation.
    \OURS outperforms prior art in RGBD classification and segmentation.
    }
    \label{tab:sota-finetune-nyu}
\end{table}

We ablate some of \OURS's key design choices in~\cref{tab:ablations}.
Together, the results suggest \OURS's performance is relatively stable under different design choices. For a faster turnaround time in the ablations, we train the model for 300 epochs.

\par \noindent \textbf{Training from scratch or finetuning.}
We compare training \OURS models from scratch on different modalities (top row) with initializing the model via image classification followed by finetuning on all modalities (second row).
For the finetuning result, we initialize \OURS (\swinB) using a pretrained \imnetFull model followed by joint finetuning on \imnetShort, \kineticsShort, and \sunrgbdShort for 100 epochs.
The model trained from scratch performs better in both image and video classification.

\par \noindent \textbf{Data ratio.}
Since the \imnetShort and \kineticsShort datasets are much larger than \sunrgbdShort, we replicate \sunrgbdShort when training \OURS.
Although replication helps, a higher replication factor hurts the model performance on \sunrgbdShort (which hints at overfitting), whereas the performance on \imnetShort and \kineticsShort is unchanged.
Based on the same logic, we undersample the \imnetFullShort dataset to have a similar size as \imnetShort. Increasing the proportion of \imnetFullShort has no effect on \imnetShort, decreases performance on \kineticsShort, and improves performance on \sunrgbdShort. Hence, we use the 0.1:1:1:10 setting for our final model.

\par \noindent \textbf{Batching strategy.}
We evaluate the two different batching strategies described in~\cref{sec:approach}, and observe that they perform similarly.
We also find that the separate batching strategy (which alternates between datasets during training) does not lead to instabilities during training.
Additionally, since it is easier to implement, we use it to train \OURS.

\par \noindent \textbf{Patch embedding model for depth channel.}
\OURS uses a separate linear+LN layer for the depth channel in RGBD images.
We compare this to using a four-channel convolutional model to embed depth patches instead, and find that the separate layer leads to better performance on \sunrgbdShort.
We also observed that using the separate layer helps \OURS transfer better to downstream RGBD tasks.%

\begin{table}[b]
    \centering
    \resizebox{\columnwidth}{!}{
    \begin{tabular}{l l | c c c}
    \rowcolor{white} & & {\color{ImageDark}\bf \imnetShort} & {\color{VideoDark}\bf \kineticsShort} &
    {\color{ThreeDDark} \bf \sunrgbdShort} \\
    \midrule
    \bf Baseline &  & 85.2 & 83.2 & 65.5 \\ %
    \hline
    \bf Finetuned &  & --0.7 & --0.9 & +0.9 \\ %
    \textbf{Data ratio}  & 0.1:1:1:1 & --0.1 & +0.3 & --0.7 \\ %
    ~\small{\imnetFullShort:\imnetShort:\kineticsShort:\sunrgbdShort}  & 0.1:1:1:10 & {\color{GrayNumber}+0} & +0.1 & +0.6 \\ %
            & 0.1:1:1:20 & {\color{GrayNumber}+0} & +0.2 & +0.6 \\ %
            & 0.1:1:1:100 & --0.1 & --0.1 & --2.1 \\ %
            & 0.3:1:1:50 & +0.1 & --1.3 & +1.5 \\ %
            & 0.6:1:1:50 & --0.2 & --3.1 & +1.0 \\ %
            & 1.0:1:1:50 & --0.1 & --4.5 & +2.0 \\ %
    \textbf{Batching} & Mixed & --0.2 & --0.1 & --0.4 \\ %
    \textbf{Patch embedding} & RGBD Conv. & --0.1 & +0.1 & --2.2 \\ %
    \end{tabular}%
    }
    \caption{{\bf Ablation study of design choices} made when training \OURS.
    Our baseline settings use a data ratio of 0.1:1:1:50, 
    the separate batching strategy, linear layers for embedding RGB and depth channels, and 300 epoch training.
    \OURS's performance is robust under different decisions.
    \OURS trained from scratch (top row) performs slightly better than a jointly finetuned model (second row).
    }
    \label{tab:ablations}
\end{table}

\section{Cross-Modal Generalization}
\label{sec:generalization}
A key advantage of \OURS over modality-specific models is that it can generalize across visual modalities.
This generalization emerges naturally because we use the same model for all modalities.
Our model is neither trained with corresponding data across modalities nor with any cross-modal consistency losses.

\begin{figure}
    \centering
    \setlength{\tabcolsep}{1pt}
    \resizebox{\linewidth}{!}{
    \begin{tabular}{l@{\hskip 0.05in}|@{\hskip 0.05in}l}
        \textbf{Query} & \multicolumn{1}{l}{\textbf{Retrieved depth maps} $\xrightarrow{\hspace*{2cm}}$} \\
        \includegraphics[height=0.15\linewidth]{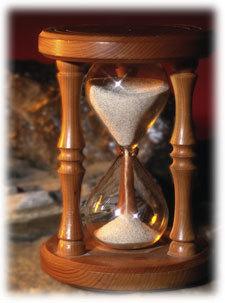} &
        \includegraphics[height=0.15\linewidth]{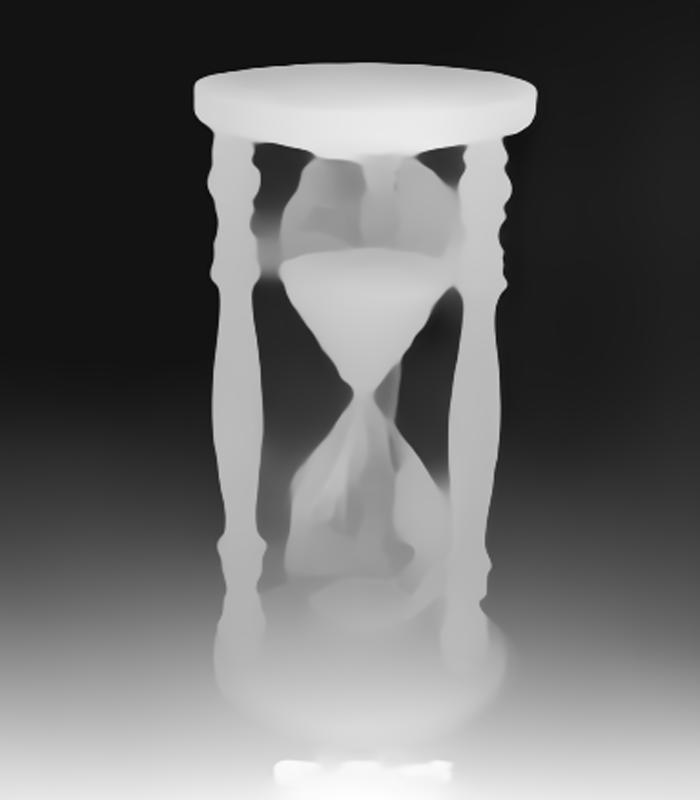} \hfill
        \includegraphics[height=0.15\linewidth]{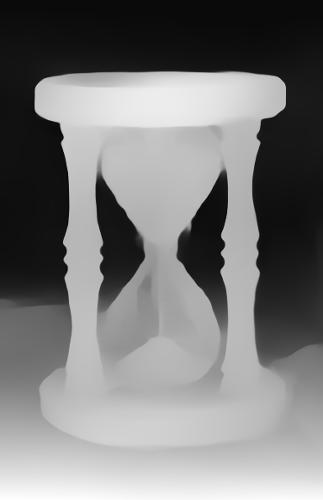} \hfill
        \includegraphics[height=0.15\linewidth]{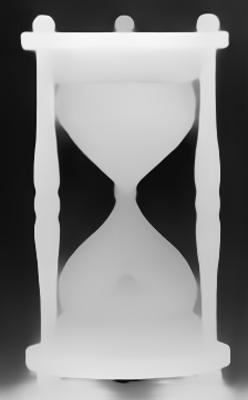} \hfill
        \includegraphics[height=0.15\linewidth]{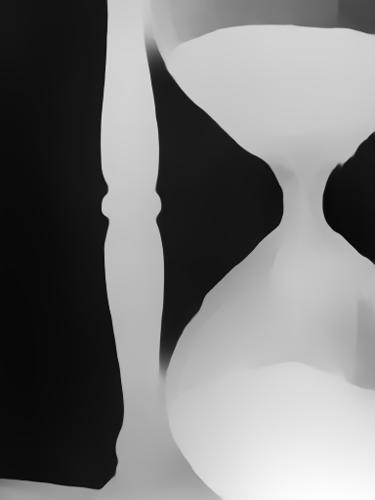} \hfill
        \includegraphics[height=0.15\linewidth]{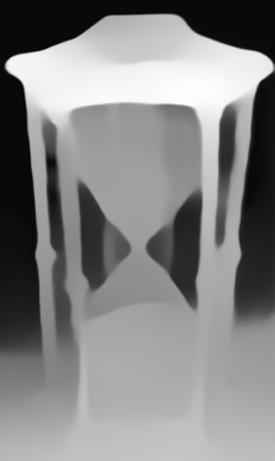} \hfill
        \includegraphics[height=0.15\linewidth]{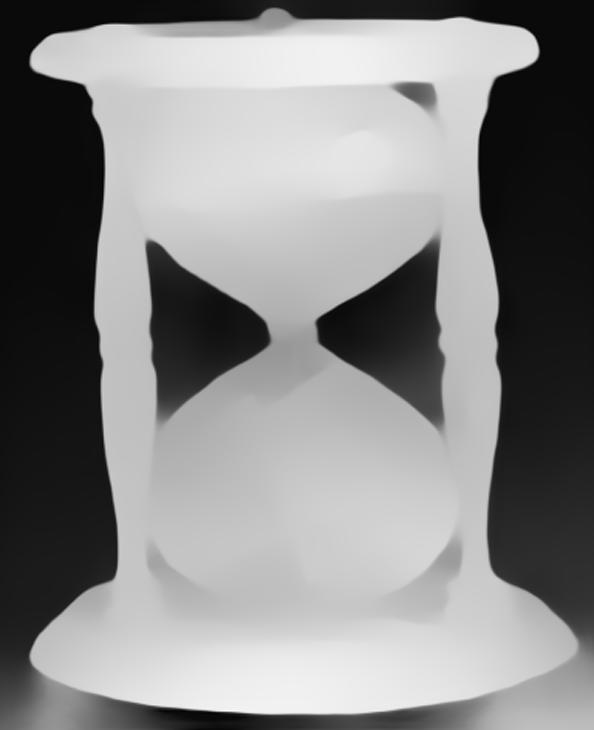} \hfill
        \includegraphics[height=0.15\linewidth]{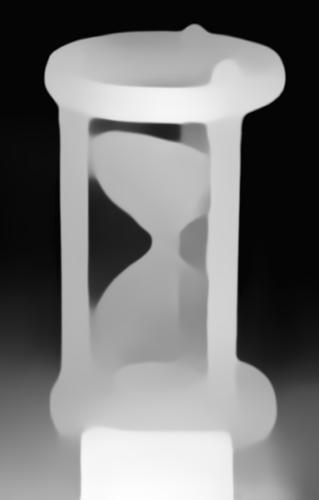} \hfill
        \includegraphics[height=0.15\linewidth]{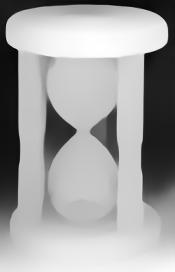} \hfill
        \\
        \includegraphics[height=0.15\linewidth]{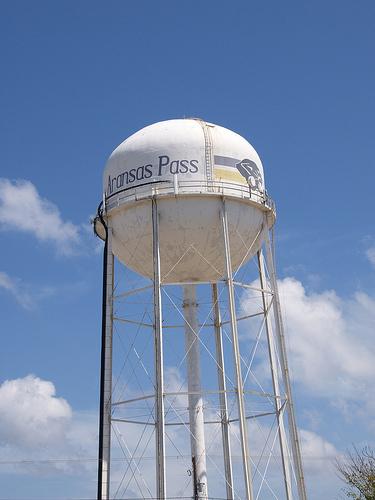} &
        \includegraphics[height=0.15\linewidth]{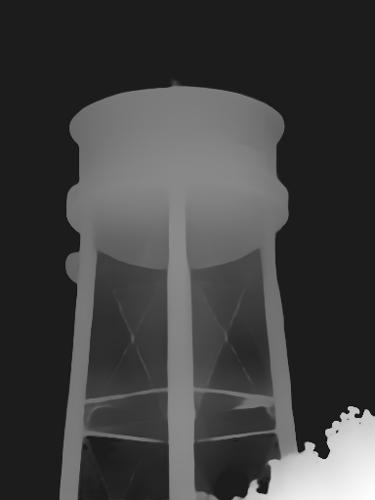} \hfill
        \includegraphics[height=0.15\linewidth]{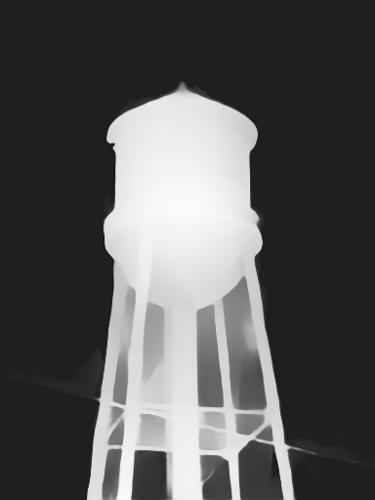} \hfill
        \includegraphics[height=0.15\linewidth]{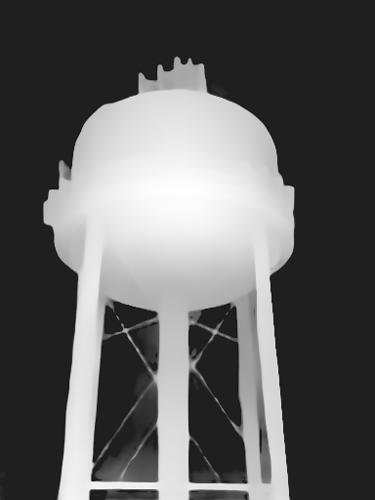} \hfill
        \includegraphics[height=0.15\linewidth]{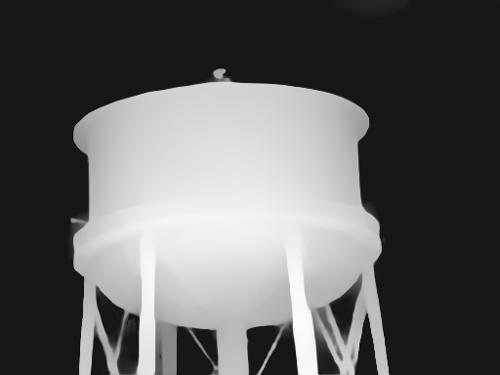} \hfill
        \includegraphics[height=0.15\linewidth]{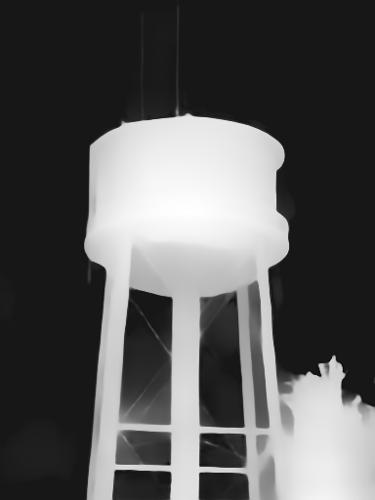} \hfill
        \includegraphics[height=0.15\linewidth]{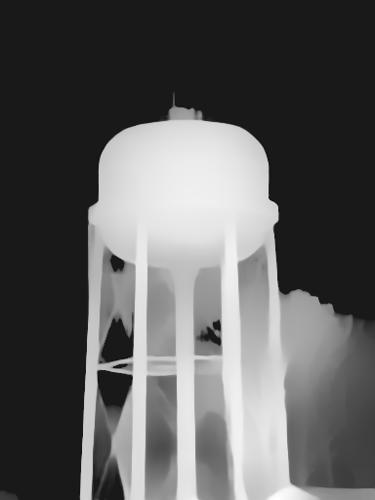} \hfill
        \includegraphics[height=0.15\linewidth]{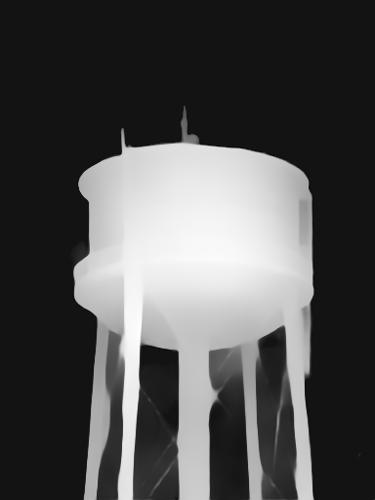} \hfill
        \\
        \includegraphics[height=0.15\linewidth]{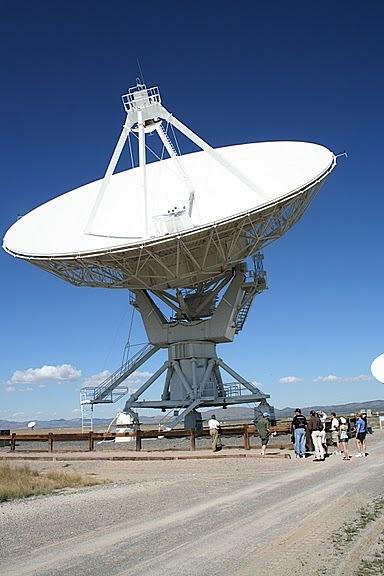} &
        \includegraphics[height=0.15\linewidth]{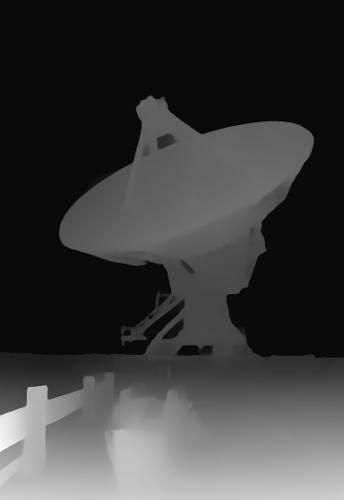} \hfill
        \includegraphics[height=0.15\linewidth]{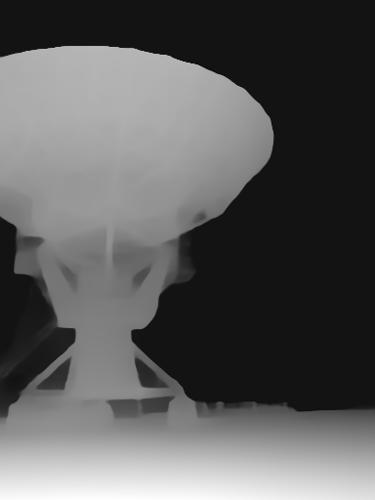} \hfill
        \includegraphics[height=0.15\linewidth]{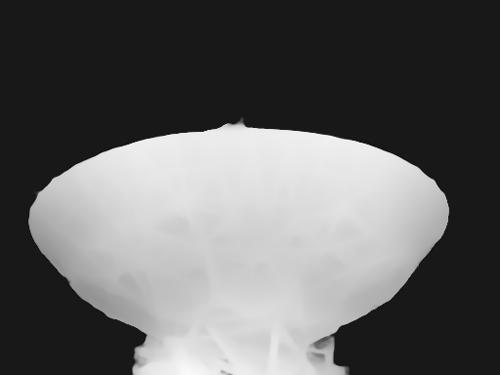} \hfill
        \includegraphics[height=0.15\linewidth]{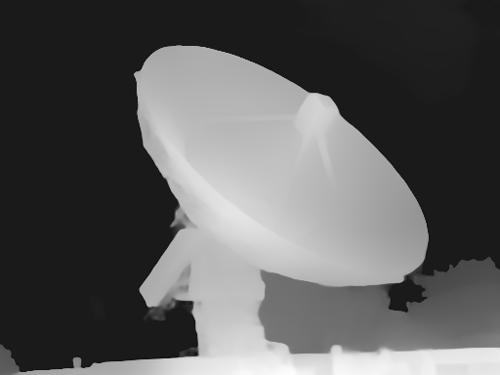} \hfill
        \includegraphics[height=0.15\linewidth]{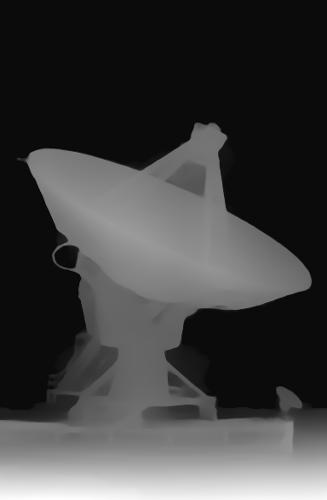} \hfill
        \includegraphics[height=0.15\linewidth]{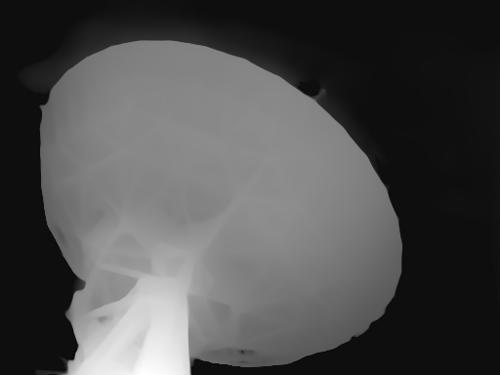} \hfill
        \\
        \includegraphics[height=0.15\linewidth]{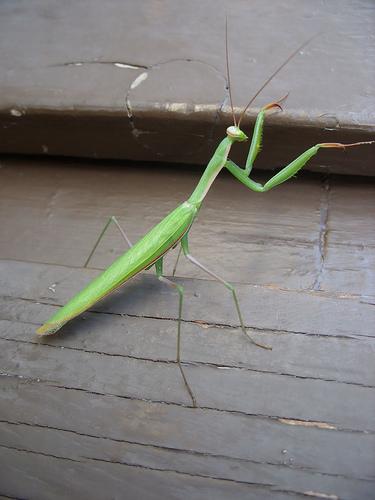} &
        \includegraphics[height=0.15\linewidth]{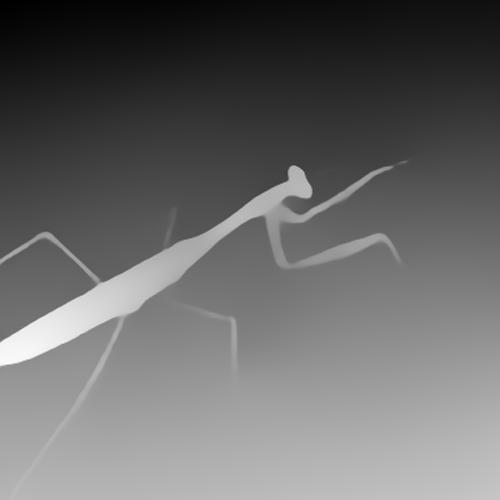} \hfill
        \includegraphics[height=0.15\linewidth]{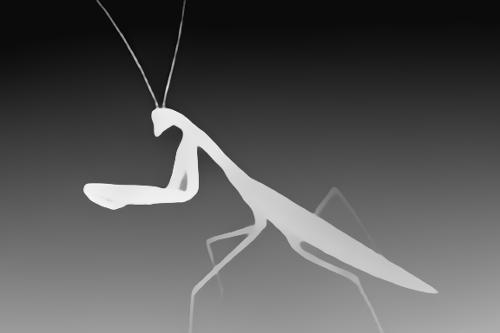} \hfill
        \includegraphics[height=0.15\linewidth]{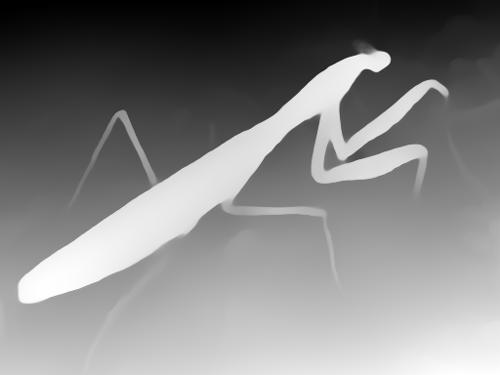} \hfill
        \includegraphics[height=0.15\linewidth]{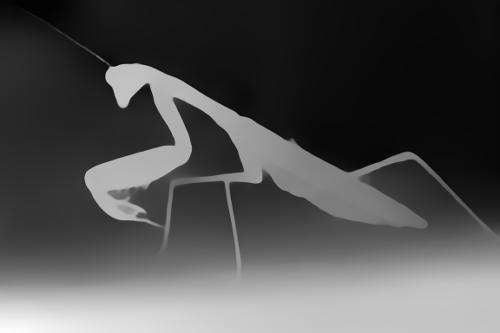} \hfill
        \includegraphics[height=0.15\linewidth]{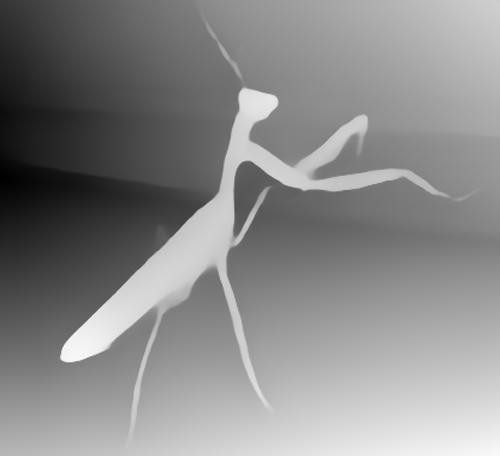} \\
    \end{tabular}%
    }
    \vspace{-0.1in}
    \caption{
        \textbf{Retrieving depth maps given RGB images} on the \imnet dataset.
        We show retrieved depth maps from the \imnetShort training set (right) for RGB image queries from the \imnetShort validation set (left).
        Although \OURS was not trained on \imnetShort depth maps, the shared visual representation enables it to retrieve depth maps that are semantically similar to the query.
    }
    \label{fig:knn_rgb_depth}
\end{figure}

\noindent \textbf{Retrieval across images and depth.}
We use the \OURS representation to retrieve depth maps given an RGB image.
To create a database of depth maps, we run a monocular depth-prediction model~\cite{ranftl2020towards} on the \imnet train set.
We note that \OURS was not trained on \imnet depth maps nor on predicted depth.
We use the \imnet val set (RGB) images as queries.
\cref{fig:knn_rgb_depth} shows five examples of retrieved maps.
These results illustrate that \OURS constructs good depth-map representations, even though it had not previously observed \imnet depth maps during training.
We emphasize that this cross-modal generalization ability is not the result of explicitly learning correspondences between visual modalities~\cite{gupta2013perceptual,simonyan2014two}.
Instead, it emerges due to the use of an almost entirely shared encoder for those modalities.

\noindent \textbf{Classifying based on different modalities.}

\begin{center}
    \setlength{\tabcolsep}{3pt}
    \begin{tabular}{l | c c c}
        \bf Method & \bf RGB & \bf D & \bf RGBD \\
        \hline
        \OURS (\swinB) & 84.3 & 63.1 & 83.7 \\ %
    \end{tabular}
 \end{center}
To quantitatively measure \OURS's generalization performance across different modalities, we perform $k$-nearest neighbor ($k$-NN, $k\!=\!20$) classification experiments on the \imnet dataset using the predicted depth maps.
We extract \OURS representations from the RGB images on the val set and measure the model's ability to retrieve images, RGBD images, and depth-only images from the train set.
We observe that \OURS produces a representation that allows for successful $k$-NN classification, which demonstrates its strong generalization performance.
Surprisingly, we observe a high accuracy is attained even when retrieving depth-images, which provide less information about the object class than RGB images.

\noindent \textbf{Retrieval across all modalities.}
We further probe the \OURS visual representations in retrieval experiments across images, videos, and depth maps.
We use the RGB images from the \imnet val set as queries and use them to retrieve similar depth maps from \imnet (predicted depth) and videos from \kinetics.
\cref{fig:teaser} %
shows examples of the resulting retrievals.
The results illustrate how \OURS supports retrieval of visual concepts across images (RGB), single-view 3D (RGBD), and videos (RGBT) using its shared representation space. %

\begin{figure}
    \centering
    \begin{tikzpicture}
        \begin{axis} [
            xlabel=Clip length (number of frames),
            ylabel={Top-1 accuracy},
            xmax=32,
            ymin=48, ymax=85,
            axis x line*=bottom,
            axis y line*=left,
            legend pos=south east,
            xmode=log,
            log ticks with fixed point,
            xtick={1,2,4,8,16,32},
            height=1.5in,
            width=0.8\linewidth,
            legend style={cells={align=left}, font=\small},
            label style={font=\small},
            tick label style={font=\small},
        ]

        \addplot[mark=*,thick,gray] plot coordinates {
            (1, 50.8)
            (2, 73.8)
            (4, 77.0)
            (8, 79.3)
            (16, 81.3)
            (32, 82.4)
        };
        \addlegendentry{Video\swinB}

        \addplot[color=VideoDark,thick,mark=square*] plot coordinates {
            (1, 69.1)
            (2, 76.6)
            (4, 78.9)
            (8, 81.7)
            (16, 83.1)
            (32, 83.8)
        };
        \addlegendentry{\OURS}
        \end{axis}
    \end{tikzpicture}
    \vspace{-0.1in}
    \caption{\textbf{Accuracy as a function of clip length} on the \kineticsShort dataset.
    Models are trained on 32-frame clips but evaluated on clips of different length (with the same fps used for frame sampling).
    The performance of \OURS degrades more gracefully than that of the Video\swinB model, and is still effective when doing frame-level inference (\emph{i.e.}, when the clip length is 1).
    }\label{fig:vid_perf_over_clip_len}
\end{figure}
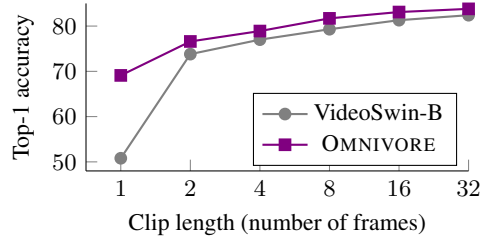

\noindent \textbf{Bridging frame-based and clip-based video models.}
\OURS's cross-modality generalization capabilities also make it more robust to changes in lengths of videos to be classified.
We demonstrate this in in~\cref{fig:vid_perf_over_clip_len},
where we classify videos using different length clips at inference time.
The model is trained with 32 frames at stride 2, and by default uses 4 clips of the same length and stride to cover the full 10 second video at inference time. In this experiment, we vary the clip length from 1 to 32, increasing the number of clips proportionally to still cover the full video in each case.
The results show that \OURS's performance degrades more gracefully as the video length decreases.
Notably, \OURS outperforms the baseline by 18.5\% at a clip length of 1 frame (frame-level inference).
This suggests that joint training on images and videos enables the model to use both temporal and spatial cues effectively.

\section{Discussion and Limitations}
\label{sec:discussion}
Although \OURS presents an advance over traditional modality-specific models, it has several limitations.
Current implementation of \OURS only works on single-view 3D images and does not generalize to other 3D representations such as voxels, point clouds, \etc
A simple approach to deal with such inputs may be to render multiple single-view 3D images from such inputs and average our predictions over those images, but such an approach would not effectively leverage multi-view information.
Another caveat is that depth inputs are not scale-invariant; we used normalizations to alleviate this issue~\cite{ranftl2020towards}.
Also, \OURS focuses only on visual modalities, so co-occurring modalities such as audio are not used.
\OURS was pretrained using only classification; using structured prediction tasks such as segmentation might yield richer representations.
We leave such extensions to future work.

\par \noindent \textbf{Ethical Considerations.}
Our study focuses on technical innovations in training models for visual recognition.
These innovations themselves appear to be neutral from an ethics point-of-view.
However, all ethical considerations that apply to other visual-recognition models apply equally to \OURS.
Any real-world deployment of a model like \OURS is best preceded by a careful analysis of that model for ethical problems, including but not limited to: performance disparities between different user groups, associations that may be harmful to some users, and predictions that may propagate stereotypes.

{\small
\bibliographystyle{ieee_fullname}
\bibliography{refs}

\begin{thebibliography}{100}\itemsep=-1pt

\bibitem{akbari2021vatt}
Hassan Akbari, Linagzhe Yuan, Rui Qian, Wei-Hong Chuang, Shih-Fu Chang, Yin
  Cui, and Boqing Gong.
\newblock Vatt: Transformers for multimodal self-supervised learning from raw
  video, audio and text.
\newblock {\em arXiv preprint arXiv:2104.11178}, 2021.

\bibitem{alayrac2020self}
Jean-Baptiste Alayrac, Adria Recasens, Rosalia Schneider, Relja Arandjelovic,
  Jason Ramapuram, Jeffrey De~Fauw, Lucas Smaira, Sander Dieleman, and Andrew
  Zisserman.
\newblock Self-supervised multimodal versatile networks.
\newblock {\em NeurIPS}, 2020.

\bibitem{arandjelovic2017look}
Relja Arandjelovic and Andrew Zisserman.
\newblock Look, listen and learn.
\newblock In {\em ICCV}, 2017.

\bibitem{arandjelovic2018objects}
Relja Arandjelovic and Andrew Zisserman.
\newblock Objects that sound.
\newblock In {\em ECCV}, 2018.

\bibitem{arnab2021vivit}
Anurag Arnab, Mostafa Dehghani, Georg Heigold, Chen Sun, Mario Lucic, and
  Cordelia Schmid.
\newblock {ViViT}: {A} video vision transformer.
\newblock In {\em ICCV}, 2021.

\bibitem{ba2016layer}
Jimmy~Lei Ba, Jamie~Ryan Kiros, and Geoffrey~E Hinton.
\newblock Layer normalization.
\newblock {\em arXiv preprint arXiv:1607.06450}, 2016.

\bibitem{bain2021frozen}
Max Bain, Arsha Nagrani, G{\"u}l Varol, and Andrew Zisserman.
\newblock Frozen in time: A joint video and image encoder for end-to-end
  retrieval.
\newblock {\em arXiv preprint arXiv:2104.00650}, 2021.

\bibitem{bertasius2021is}
Gedas Bertasius, Heng Wang, and Lorenzo Torresani.
\newblock Is space-time attention all you need for video understanding?
\newblock In {\em ICML}, 2021.

\bibitem{caglayan2020CNNrandRNN}
Ali Caglayan, Nevrez Imamoglu, Ahmet~Burak Can, and Ryosuke Nakamura.
\newblock When cnns meet random rnns: Towards multi-level analysis for rgb-d
  object and scene recognition.
\newblock {\em arXiv preprint arXiv:2004.12349}, 2020.

\bibitem{cao2021shapeconv}
Jinming Cao, Hanchao Leng, Dani Lischinski, Danny Cohen-Or, Changhe Tu, and
  Yangyan Li.
\newblock Shapeconv: Shape-aware convolutional layer for indoor rgb-d semantic
  segmentation.
\newblock In {\em ICCV}, 2021.

\bibitem{carion2020end}
Nicolas Carion, Francisco Massa, Gabriel Synnaeve, Nicolas Usunier, Alexander
  Kirillov, and Sergey Zagoruyko.
\newblock End-to-end object detection with transformers.
\newblock In {\em ECCV}, 2020.

\bibitem{caron2021emerging}
Mathilde Caron, Hugo Touvron, Ishan Misra, Herv\'e J\'egou, Julien Mairal,
  Piotr Bojanowski, and Armand Joulin.
\newblock Emerging properties in self-supervised vision transformers.
\newblock In {\em ICCV}, 2021.

\bibitem{carreira2017quo}
Jo\~ao Carreira and Andrew Zisserman.
\newblock Quo vadis, action recognition? a new model and the kinetics dataset.
\newblock In {\em CVPR}, 2017.

\bibitem{caruana1997}
Rich Caruana.
\newblock Multitask learning.
\newblock {\em Machine Learning}, 1997.

\bibitem{castrejon2016learning}
Lluis Castrejon, Yusuf Aytar, Carl Vondrick, Hamed Pirsiavash, and Antonio
  Torralba.
\newblock Learning aligned cross-modal representations from weakly aligned
  data.
\newblock In {\em CVPR}, 2016.

\bibitem{chen2020-SAGate}
Xiaokang Chen, Kwan-Yee Lin, Jingbo Wang, Wayne Wu, Chen Qian, Hongsheng Li,
  and Gang Zeng.
\newblock Bi-directional cross-modality feature propagation with
  separation-and-aggregation gate for rgb-d semantic segmentation.
\newblock In {\em ECCV}, 2020.

\bibitem{chen2020uniter}
Yen-Chun Chen, Linjie Li, Licheng Yu, Ahmed El~Kholy, Faisal Ahmed, Zhe Gan, Yu
  Cheng, and Jingjing Liu.
\newblock Uniter: Universal image-text representation learning.
\newblock In {\em ECCV}, 2020.

\bibitem{choy20194d}
Christopher Choy, JunYoung Gwak, and Silvio Savarese.
\newblock 4d spatio-temporal convnets: Minkowski convolutional neural networks.
\newblock In {\em CVPR}, 2019.

\bibitem{cubuk2020randaugment}
Ekin~D Cubuk, Barret Zoph, Jonathon Shlens, and Quoc~V Le.
\newblock Randaugment: Practical automated data augmentation with a reduced
  search space.
\newblock In {\em CVPR}, 2020.

\bibitem{damen2021rescaling}
Dima Damen, Hazel Doughty, Giovanni~Maria Farinella, Antonino Furnari,
  Evangelos Kazakos, Jian Ma, Davide Moltisanti, Jonathan Munro, Toby Perrett,
  Will Price, et~al.
\newblock Rescaling egocentric vision.
\newblock {\em IJCV}, 2021.

\bibitem{dosovitskiy2020image}
Alexey Dosovitskiy, Lucas Beyer, Alexander Kolesnikov, Dirk Weissenborn,
  Xiaohua Zhai, Thomas Unterthiner, Mostafa Dehghani, Matthias Minderer, Georg
  Heigold, Sylvain Gelly, et~al.
\newblock An image is worth 16x16 words: Transformers for image recognition at
  scale.
\newblock In {\em ICLR}, 2021.

\bibitem{du2019translate}
Dapeng Du, Limin Wang, Huiling Wang, Kai Zhao, and Gangshan Wu.
\newblock Translate-to-recognize networks for rgb-d scene recognition.
\newblock In {\em CVPR}, 2019.

\bibitem{eigen2015predicting}
David Eigen and Rob Fergus.
\newblock Predicting depth, surface normals and semantic labels with a common
  multi-scale convolutional architecture.
\newblock In {\em ICCV}, 2015.

\bibitem{fan2021multiscale}
Haoqi Fan, Bo Xiong, Karttikeya Mangalam, Yanghao Li, Zhicheng Yan, Jitendra
  Malik, and Christoph Feichtenhofer.
\newblock Multiscale vision transformers.
\newblock In {\em ICCV}, 2021.

\bibitem{feichtenhofer2019slowfast}
Christoph Feichtenhofer, Haoqi Fan, Jitendra Malik, and Kaiming He.
\newblock Slowfast networks for video recognition.
\newblock In {\em ICCV}, 2019.

\bibitem{fukushima1980self}
Kunihiko Fukushima.
\newblock A self-organizing neural network model for a mechanism of pattern
  recognition unaffected by shift in position.
\newblock {\em Biol. Cybern.}, 1980.

\bibitem{ghiasi2021multi}
Golnaz Ghiasi, Barret Zoph, Ekin~D Cubuk, Quoc~V Le, and Tsung-Yi Lin.
\newblock Multi-task self-training for learning general representations.
\newblock In {\em ICCV}, 2021.

\bibitem{girdhar2019video}
Rohit Girdhar, Jo\~ao Carreira, Carl Doersch, and Andrew Zisserman.
\newblock Video action transformer network.
\newblock In {\em CVPR}, 2019.

\bibitem{girdhar2021anticipative}
Rohit Girdhar and Kristen Grauman.
\newblock {Anticipative Video Transformer}.
\newblock In {\em ICCV}, 2021.

\bibitem{gong2014improving}
Yunchao Gong, Liwei Wang, Micah Hodosh, Julia Hockenmaier, and Svetlana
  Lazebnik.
\newblock Improving image-sentence embeddings using large weakly annotated
  photo collections.
\newblock In {\em ECCV}, 2014.

\bibitem{goyal2017something}
Raghav Goyal, Samira Ebrahimi~Kahou, Vincent Michalski, Joanna Materzynska,
  Susanne Westphal, Heuna Kim, Valentin Haenel, Ingo Fruend, Peter Yianilos,
  Moritz Mueller-Freitag, Florian Hoppe, Christian Thurau, Ingo Bax, and Roland
  Memisevic.
\newblock The ``something something'' video database for learning and
  evaluating visual common sense.
\newblock In {\em ICCV}, 2017.

\bibitem{graham20183d}
Benjamin Graham, Martin Engelcke, and Laurens van~der Maaten.
\newblock 3d semantic segmentation with submanifold sparse convolutional
  networks.
\newblock In {\em CVPR}, 2018.

\bibitem{gupta2013perceptual}
Saurabh Gupta, Pablo Arbelaez, and Jitendra Malik.
\newblock Perceptual organization and recognition of indoor scenes from rgb-d
  images.
\newblock In {\em CVPR}, 2013.

\bibitem{he2016deep}
Kaiming He, Xiangyu Zhang, Shaoqing Ren, and Jian Sun.
\newblock Deep residual learning for image recognition.
\newblock In {\em CVPR}, 2016.

\bibitem{herzig2021object}
Roei Herzig, Elad Ben-Avraham, Karttikeya Mangalam, Amir Bar, Gal Chechik, Anna
  Rohrbach, Trevor Darrell, and Amir Globerson.
\newblock Object-region video transformers.
\newblock {\em arXiv preprint arXiv:2110.06915}, 2021.

\bibitem{iNaturalist}
Grant~Van Horn, Oisin~Mac Aodha, Yang Song, Yin Cui, Chen Sun, Alex Shepard,
  Hartwig Adam, Pietro Perona, and Serge Belongie.
\newblock The inaturalist species classification and detection dataset.
\newblock In {\em CVPR}, 2018.

\bibitem{hu2021unit}
Ronghang Hu and Amanpreet Singh.
\newblock Unit: Multimodal multitask learning with a unified transformer.
\newblock In {\em ICCV}, 2021.

\bibitem{jaegle2021perceiver}
Andrew Jaegle, Felix Gimeno, Andrew Brock, Andrew Zisserman, Oriol Vinyals, and
  Joao Carreira.
\newblock Perceiver: General perception with iterative attention.
\newblock {\em ICML}, 2021.

\bibitem{kaiser2017one}
Lukasz Kaiser, Aidan~N Gomez, Noam Shazeer, Ashish Vaswani, Niki Parmar, Llion
  Jones, and Jakob Uszkoreit.
\newblock One model to learn them all.
\newblock {\em arXiv preprint arXiv:1706.05137}, 2017.

\bibitem{kamath2021mdetr}
Aishwarya Kamath, Mannat Singh, Yann LeCun, Gabriel Synnaeve, Ishan Misra, and
  Nicolas Carion.
\newblock Mdetr-modulated detection for end-to-end multi-modal understanding.
\newblock In {\em ICCV}, 2021.

\bibitem{karpathy2015deep}
Andrej Karpathy and Li Fei-Fei.
\newblock Deep visual-semantic alignments for generating image descriptions.
\newblock In {\em CVPR}, 2015.

\bibitem{kay2017kinetics}
Will Kay, Joao Carreira, Karen Simonyan, Brian Zhang, Chloe Hillier, Sudheendra
  Vijayanarasimhan, Fabio Viola, Tim Green, Trevor Back, Paul Natsev, et~al.
\newblock The kinetics human action video dataset.
\newblock {\em arXiv preprint arXiv:1705.06950}, 2017.

\bibitem{kazakos2021MTCN}
Evangelos Kazakos, Jaesung Huh, Arsha Nagrani, Andrew Zisserman, and Dima
  Damen.
\newblock With a little help from my temporal context: Multimodal egocentric
  action recognition.
\newblock In {\em BMVC}, 2021.

\bibitem{kokkinos2017ubernet}
Iasonas Kokkinos.
\newblock Uber{N}et: Training a universal convolutional neural network for
  low-, mid-, and high-level vision using diverse datasets and limited memory.
\newblock In {\em CVPR}, 2017.

\bibitem{komkov2020mutual}
Stepan Komkov, Maksim Dzabraev, and Aleksandr Petiushko.
\newblock Mutual modality learning for video action classification.
\newblock {\em arXiv preprint arXiv:2011.02543}, 2020.

\bibitem{krizhevsky2012imagenet}
Alex Krizhevsky, Ilya Sutskever, and Geoffrey~E Hinton.
\newblock Imagenet classification with deep convolutional neural networks.
\newblock {\em NeurIPS}, 2012.

\bibitem{laptev2003space}
Ivan Laptev and Tony Lindeberg.
\newblock Space-time interest points.
\newblock In {\em ICCV}, 2003.

\bibitem{lecun1998gradient}
Yann LeCun, L{\'e}on Bottou, Yoshua Bengio, and Patrick Haffner.
\newblock Gradient-based learning applied to document recognition.
\newblock {\em Proceedings of the IEEE}, 1998.

\bibitem{li2020oscar}
Xiujun Li, Xi Yin, Chunyuan Li, Pengchuan Zhang, Xiaowei Hu, Lei Zhang, Lijuan
  Wang, Houdong Hu, Li Dong, Furu Wei, Yejin Choi, and Jianfeng Gao.
\newblock Oscar: Object-semantics aligned pre-training for vision-language
  tasks.
\newblock In {\em ECCV}, 2020.

\bibitem{li2018df2net}
Yabei Li, Junge Zhang, Yanhua Cheng, Kaiqi Huang, and Tieniu Tan.
\newblock Df\({}^{\mbox{2}}\)net: Discriminative feature learning and fusion
  network for {RGB-D} indoor scene classification.
\newblock In {\em AAAI}, 2018.

\bibitem{liu2021swin}
Ze Liu, Yutong Lin, Yue Cao, Han Hu, Yixuan Wei, Zheng Zhang, Stephen Lin, and
  Baining Guo.
\newblock Swin transformer: Hierarchical vision transformer using shifted
  windows.
\newblock In {\em ICCV}, 2021.

\bibitem{liu2021video}
Ze Liu, Jia Ning, Yue Cao, Yixuan Wei, Zheng Zhang, Stephen Lin, and Han Hu.
\newblock Video swin transformer.
\newblock {\em arXiv preprint arXiv:2106.13230}, 2021.

\bibitem{loshchilov2017decoupled}
Ilya Loshchilov and Frank Hutter.
\newblock Decoupled weight decay regularization.
\newblock {\em arXiv preprint arXiv:1711.05101}, 2017.

\bibitem{loshchilov2016sgdr}
Ilya Loshchilov and Frank Hutter.
\newblock Sgdr: Stochastic gradient descent with warm restarts.
\newblock In {\em ICLR}, 2017.

\bibitem{lowe2004distinctive}
David~G Lowe.
\newblock Distinctive image features from scale-invariant keypoints.
\newblock {\em IJCV}, 2004.

\bibitem{lu2019vilbert}
Jiasen Lu, Dhruv Batra, Devi Parikh, and Stefan Lee.
\newblock Vilbert: Pretraining task-agnostic visiolinguistic representations
  for vision-and-language tasks.
\newblock {\em arXiv preprint arXiv:1908.02265}, 2019.

\bibitem{lu202012}
Jiasen Lu, Vedanuj Goswami, Marcus Rohrbach, Devi Parikh, and Stefan Lee.
\newblock 12-in-1: Multi-task vision and language representation learning.
\newblock In {\em CVPR}, 2020.

\bibitem{maninis2019attentive}
Kevis-Kokitsi Maninis, Ilija Radosavovic, and Iasonas Kokkinos.
\newblock Attentive single-tasking of multiple tasks.
\newblock In {\em CVPR}, 2019.

\bibitem{miech2020end}
Antoine Miech, Jean-Baptiste Alayrac, Lucas Smaira, Ivan Laptev, Josef Sivic,
  and Andrew Zisserman.
\newblock End-to-end learning of visual representations from uncurated
  instructional videos.
\newblock In {\em CVPR}, 2020.

\bibitem{misra2021-3detr}
Ishan Misra, Rohit Girdhar, and Armand Joulin.
\newblock {An End-to-End Transformer Model for 3D Object Detection}.
\newblock In {\em ICCV}, 2021.

\bibitem{misra2016cross}
Ishan Misra, Abhinav Shrivastava, Abhinav Gupta, and Martial Hebert.
\newblock Cross-stitch networks for multi-task learning.
\newblock In {\em CVPR}, 2016.

\bibitem{morgado2021robust}
Pedro Morgado, Ishan Misra, and Nuno Vasconcelos.
\newblock Robust audio-visual instance discrimination.
\newblock In {\em CVPR}, 2021.

\bibitem{morgado2021audio}
Pedro Morgado, Nuno Vasconcelos, and Ishan Misra.
\newblock Audio-visual instance discrimination with cross-modal agreement.
\newblock In {\em CVPR}, 2021.

\bibitem{nagrani2021attention}
Arsha Nagrani, Shan Yang, Anurag Arnab, Aren Jansen, Cordelia Schmid, and Chen
  Sun.
\newblock Attention bottlenecks for multimodal fusion.
\newblock In {\em NeurIPS}, 2021.

\bibitem{Silberman_ECCV12_NYUv2}
Pushmeet~Kohli Nathan~Silberman, Derek~Hoiem and Rob Fergus.
\newblock Indoor segmentation and support inference from rgbd images.
\newblock In {\em ECCV}, 2012.

\bibitem{neimark2021video}
Daniel Neimark, Omri Bar, Maya Zohar, and Dotan Asselmann.
\newblock Video transformer network.
\newblock {\em arXiv preprint arXiv:2102.00719}, 2021.

\bibitem{owens}
Andrew Owens and Alexei~A Efros.
\newblock Audio-visual scene analysis with self-supervised multisensory
  features.
\newblock In {\em ECCV}, 2018.

\bibitem{pan20213d}
Xuran Pan, Zhuofan Xia, Shiji Song, Li~Erran Li, and Gao Huang.
\newblock 3d object detection with pointformer.
\newblock In {\em CVPR}, 2021.

\bibitem{parkhi2012cats}
Omkar~M Parkhi, Andrea Vedaldi, Andrew Zisserman, and CV Jawahar.
\newblock Cats and dogs.
\newblock In {\em CVPR}, 2012.

\bibitem{parmar2018image}
Niki Parmar, Ashish Vaswani, Jakob Uszkoreit, Lukasz Kaiser, Noam Shazeer,
  Alexander Ku, and Dustin Tran.
\newblock Image transformer.
\newblock In {\em ICML}, 2018.

\bibitem{patrick2020multi}
Mandela Patrick, Yuki~M Asano, Ruth Fong, Jo{\~a}o~F Henriques, Geoffrey Zweig,
  and Andrea Vedaldi.
\newblock Multi-modal self-supervision from generalized data transformations.
\newblock {\em arXiv preprint arXiv:2003.04298}, 2020.

\bibitem{patrick2021keeping}
Mandela Patrick, Dylan Campbell, Yuki~M Asano, Ishan Misra~Florian Metze,
  Christoph Feichtenhofer, Andrea Vedaldi, and Jo\~ao Henriques.
\newblock Keeping your eye on the ball: Trajectory attention in video
  transformers.
\newblock In {\em NeurIPS}, 2021.

\bibitem{polyak1992acceleration}
Boris~T Polyak and Anatoli~B Juditsky.
\newblock Acceleration of stochastic approximation by averaging.
\newblock {\em SIAM journal on control and optimization}, 1992.

\bibitem{ranftl2020towards}
Ren\'{e} Ranftl, Katrin Lasinger, David Hafner, Konrad Schindler, and Vladlen
  Koltun.
\newblock Towards robust monocular depth estimation: Mixing datasets for
  zero-shot cross-dataset transfer.
\newblock {\em TPAMI}, 2020.

\bibitem{ILSVRC15}
Olga Russakovsky, Jia Deng, Hao Su, Jonathan Krause, Sanjeev Satheesh, Sean Ma,
  Zhiheng Huang, Andrej Karpathy, Aditya Khosla, Michael Bernstein,
  Alexander~C. Berg, and Li Fei-Fei.
\newblock {ImageNet Large Scale Visual Recognition Challenge}.
\newblock {\em IJCV}, 2015.

\bibitem{sener2021technical}
Fadime Sener, Dibyadip Chatterjee, and Angela Yao.
\newblock Technical report: Temporal aggregate representations.
\newblock {\em arXiv preprint arXiv:2106.03152}, 2021.

\bibitem{simonyan2014two}
Karen Simonyan and Andrew Zisserman.
\newblock Two-stream convolutional networks for action recognition in videos.
\newblock In {\em NeurIPS}, 2014.

\bibitem{singh2022revisiting}
Mannat Singh, Laura Gustafson, Aaron Adcock, Vinicius de~Freitas Reis, Bugra
  Gedik, Raj~Prateek Kosaraju, Dhruv Mahajan, Ross Girshick, Piotr Doll{\'a}r,
  and Laurens van~der Maaten.
\newblock Revisiting weakly supervised pre-training of visual perception
  models.
\newblock In {\em CVPR}, 2022.

\bibitem{song2015sun}
Shuran Song, Samuel~P Lichtenberg, and Jianxiong Xiao.
\newblock Sun rgb-d: A rgb-d scene understanding benchmark suite.
\newblock In {\em CVPR}, 2015.

\bibitem{song2020image}
Xinhang Song, Shuqiang Jiang, Bohan Wang, Chengpeng Chen, and Gongwei Chen.
\newblock Image representations with spatial object-to-object relations for
  rgb-d scene recognition.
\newblock {\em TIP}, 2020.

\bibitem{soomro2012ucf101}
Khurram Soomro, Amir~Roshan Zamir, and Mubarak Shah.
\newblock {UCF101}: {A} dataset of 101 human action classes from videos in the
  wild.
\newblock {\em CRCV-TR-12-01}, 2012.

\bibitem{srivastava2014dropout}
Nitish Srivastava, Geoffrey Hinton, Alex Krizhevsky, Ilya Sutskever, and Ruslan
  Salakhutdinov.
\newblock Dropout: a simple way to prevent neural networks from overfitting.
\newblock {\em JMLR}, 2014.

\bibitem{su2019vl}
Weijie Su, Xizhou Zhu, Yue Cao, Bin Li, Lewei Lu, Furu Wei, and Jifeng Dai.
\newblock Vl-bert: Pre-training of generic visual-linguistic representations.
\newblock {\em arXiv preprint arXiv:1908.08530}, 2019.

\bibitem{szegedy2015going}
Christian Szegedy, Wei Liu, Yangqing Jia, Pierre Sermanet, Scott Reed, Dragomir
  Anguelov, Dumitru Erhan, Vincent Vanhoucke, and Andrew Rabinovich.
\newblock Going deeper with convolutions.
\newblock In {\em CVPR}, 2015.

\bibitem{szegedy2016rethinking}
Christian Szegedy, Vincent Vanhoucke, Sergey Ioffe, Jon Shlens, and Zbigniew
  Wojna.
\newblock Rethinking the inception architecture for computer vision.
\newblock In {\em CVPR}, 2016.

\bibitem{tan2019lxmert}
Hao Tan and Mohit Bansal.
\newblock Lxmert: Learning cross-modality encoder representations from
  transformers.
\newblock {\em arXiv preprint arXiv:1908.07490}, 2019.

\bibitem{tan2021vimpac}
Hao Tan, Jie Lei, Thomas Wolf, and Mohit Bansal.
\newblock {VIMPAC}: {V}ideo pre-training via masked token prediction and
  contrastive learning.
\newblock {\em arXiv preprint arXiv:2106.11250}, 2021.

\bibitem{touvron2021training}
Hugo Touvron, Matthieu Cord, Matthijs Douze, Francisco Massa, Alexandre
  Sablayrolles, and Herv{\'e} J{\'e}gou.
\newblock Training data-efficient image transformers \& distillation through
  attention.
\newblock In {\em ICML}, 2021.

\bibitem{touvron2019fixing}
Hugo Touvron, Andrea Vedaldi, Matthijs Douze, and Herv{\'e} J{\'e}gou.
\newblock Fixing the train-test resolution discrepancy.
\newblock In {\em NeurIPS}, 2019.

\bibitem{tran2015learning}
Du Tran, Lubomir Bourdev, Rob Fergus, Lorenzo Torresani, and Manohar Paluri.
\newblock Learning spatiotemporal features with 3d convolutional networks.
\newblock In {\em CVPR}, 2015.

\bibitem{tran2018closer}
Du Tran, Heng Wang, Lorenzo Torresani, Jamie Ray, Yann LeCun, and Manohar
  Paluri.
\newblock A closer look at spatiotemporal convolutions for action recognition.
\newblock In {\em CVPR}, 2018.

\bibitem{vaswani2017attention}
Ashish Vaswani, Noam Shazeer, Niki Parmar, Jakob Uszkoreit, Llion Jones,
  Aidan~N Gomez, Lukasz Kaiser, and Illia Polosukhin.
\newblock Attention is all you need.
\newblock In {\em NeurIPS}, 2017.

\bibitem{wang2021pyramid}
Wenhai Wang, Enze Xie, Xiang Li, Deng-Ping Fan, Kaitao Song, Ding Liang, Tong
  Lu, Ping Luo, and Ling Shao.
\newblock Pyramid vision transformer: A versatile backbone for dense prediction
  without convolutions.
\newblock In {\em ICCV}, 2021.

\bibitem{wang2018non}
Xiaolong Wang, Ross Girshick, Abhinav Gupta, and Kaiming He.
\newblock Non-local neural networks.
\newblock In {\em CVPR}, 2018.

\bibitem{xiao2018unified}
Tete Xiao, Yingcheng Liu, Bolei Zhou, Yuning Jiang, and Jian Sun.
\newblock Unified perceptual parsing for scene understanding.
\newblock In {\em ECCV}, 2018.

\bibitem{xie2020adversarial}
Cihang Xie, Mingxing Tan, Boqing Gong, Jiang Wang, Alan~L Yuille, and Quoc~V
  Le.
\newblock Adversarial examples improve image recognition.
\newblock In {\em CVPR}, 2020.

\bibitem{yue2021two}
Yuchun Yue, Wujie Zhou, Jingsheng Lei, and Lu Yu.
\newblock Two-stage cascaded decoder for semantic segmentation of rgb-d images.
\newblock {\em IEEE Signal Processing Letters}, 2021.

\bibitem{yun2019cutmix}
Sangdoo Yun, Dongyoon Han, Seong~Joon Oh, Sanghyuk Chun, Junsuk Choe, and
  Youngjoon Yoo.
\newblock Cutmix: Regularization strategy to train strong classifiers with
  localizable features.
\newblock In {\em ICCV}, 2019.

\bibitem{ZamirSSGMS18}
Amir~Roshan Zamir, Alexander Sax, William~B. Shen, Leonidas~J. Guibas, Jitendra
  Malik, and Silvio Savarese.
\newblock Taskonomy: Disentangling task transfer learning.
\newblock In {\em CVPR}, 2018.

\bibitem{zhang2021cover}
Bowen Zhang, Jiahui Yu, Christopher Fifty, Wei Han, Andrew~M. Dai, Ruoming
  Pang, and Fei Sha.
\newblock Co-training transformer with videos and images improves action
  recognition.
\newblock {\em arXiv preprint arXiv:2112.07175}, 2021.

\bibitem{zhang2017mixup}
Hongyi Zhang, Moustapha Cisse, Yann~N Dauphin, and David Lopez-Paz.
\newblock mixup: Beyond empirical risk minimization.
\newblock In {\em ICLR}, 2018.

\bibitem{zhang2014facial}
Zhanpeng Zhang, Ping Luo, Chen~Change Loy, and Xiaou Tang.
\newblock Facial landmark detection by deep multi-task learning.
\newblock In {\em ECCV}, 2014.

\bibitem{zhao2020point}
Hengshuang Zhao, Li Jiang, Jiaya Jia, Philip Torr, and Vladlen Koltun.
\newblock Point transformer.
\newblock In {\em ICCV}, 2021.

\bibitem{zhong2020random}
Zhun Zhong, Liang Zheng, Guoliang Kang, Shaozi Li, and Yi Yang.
\newblock Random erasing data augmentation.
\newblock In {\em AAAI}, 2020.

\bibitem{zhou2017places}
Bolei Zhou, Agata Lapedriza, Aditya Khosla, Aude Oliva, and Antonio Torralba.
\newblock Places: A 10 million image database for scene recognition.
\newblock {\em TPAMI}, 2017.

\end{thebibliography}
}

\clearpage

\appendix
\section{Implementation details for Pretraining}
\label{app:pretraining_details}

We train using AdamW with a batch size of 4096 for each dataset, and use a cosine learning rate (LR) schedule with linear warm up and cool down phases for the first and last 10\% of training, respectively.
We train for 500 epochs with a peak LR of $2\cdot{10}^{-3}$ and a weight decay of $5\cdot{10}^{-2}$.
\swinT, \swinS and \swinL use a window size of $8\!\times\!7\!\times\!7$, whereas \swinB uses a window size of $16\!\times\!7\!\times\!7$.
The models are trained with stochastic depth with a drop rate of $0.1$ for \swinT, 0.2 for \swinS, and 0.3 for \swinB, and \swinL.
We use exponential moving average (EMA)~\cite{polyak1992acceleration} with a decay of $10^{-4}$ and report the best results during training since EMA results peak before the end of training.

For \imnetShort and \imnetFullShort we use RandAugment~\cite{cubuk2020randaugment}, mixup~\cite{zhang2017mixup}, CutMix~\cite{yun2019cutmix}, label smoothing~\cite{szegedy2016rethinking}, and Random Erasing~\cite{zhong2020random} with the same settings as used in~\cite{touvron2021training}, and color jittering of 0.4. For \sunrgbd we clamp and normalize the disparity channel, drop the RGB channels with a probability of 0.5, and we also apply 0.5 Dropout~\cite{srivastava2014dropout} before the linear head when pre-training with \imnetFull. For \kinetics we use mixup, CutMix and label smoothing, and Dropout of $0.5$ before the linear head.

\section{Details on the Transfer Tasks}
\label{app:details_transfer}

\subsection{Image Classification}
We finetune all models on the downstream tasks for 100 epochs and optimize the models with mini-batch SGD.
We use a half-wave cosine learning rate~\cite{loshchilov2016sgdr} and set the weight decay to zero.
For all models, including the modality-specific models, we perform a grid search for the best learning rate in the range [5e-3, 1e-2, 2e-2, 4e-2, 8e-2, 1e-1, 2e-1, 3e-1, 4e-1, 5e-1, 6e-1] and drop path in [0.1, 0.3].
We use the strong augmentations from~\cite{touvron2021training} for finetuning.
For the evaluations in~\cref{tab:inflation_vs_us_finetuned,tab:sota-finetune-image}, we follow~\cite{singh2022revisiting} and resize the images to shortest side of $224$px and evaluate the models on the center crop of $224\times224$.
For higher resolution ($384$px) evaluations in~\cref{tab:sota-finetune-image}, we similarly resize the images to shortest side of $384$px and evaluate the models on the center crop of $384\times384$. We also increase the spatial window size for all the Swin models from $7$ to $12$.
\subsection{Video Classification}

In~\cref{tab:inflation_vs_us_finetuned}, we finetune video models using hyperparameters as described in~\cite{liu2021video}. For \sthsth, we finetune for 60 epochs with AdamW optimizer. We use half-wave cosine learning rate with warmup. We start the learning rate from $10^{-6}$ and linearly warmup to a peak learning rate of $6\cdot\!10^{-3}$ over 5\% of the training, and rest 95\% we use half-wave cosine schedule to decay the learning rate back to $10^{-6}$. We train the classification head with this learning rate, and the backbone with $0.1\!\times$ the above learning rate. Throughout we use a weight decay of 0.05. We use a batch size of $4\!\times 64$ distributed over 64 32GB GPUs.
For \epic, we use similar hyperparamters with only difference being that we use a peak learning rate of $2\!\cdot\!10^{-3}$ %
and we train for 150 epochs.
These settings provided better performance for the modality-specific baseline, and we use it for finetuning both the baseline and \OURS models.

\begin{figure}[!t]
    \centering
    \includegraphics[width=\linewidth]{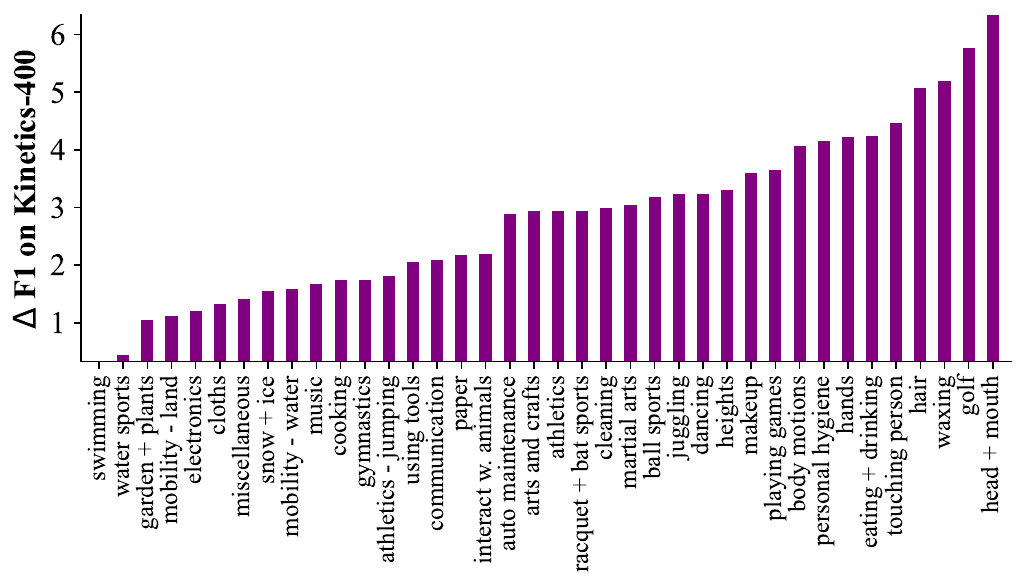}
    \vspace{-0.1in}
    \caption{\textbf{Gain of \OURS over baseline on Action recognition (per group).} We plot the gain in per-class F1-score on the \kineticsShort dataset for all the action groups defined in~\cite{carreira2017quo}.
    The baseline model is first pretrained on \imnet and then fine-tuned on \kineticsShort whereas \OURS is trained jointly on \imnet, \kineticsShort and the single-view 3D \sunrgbd dataset.
    \OURS improves the performance for all the 38 groups.
    }
    \label{fig:per_group_gain_k400_full}
\end{figure}

\begin{figure*}[!t]
    \centering
    \includegraphics[width=\linewidth]{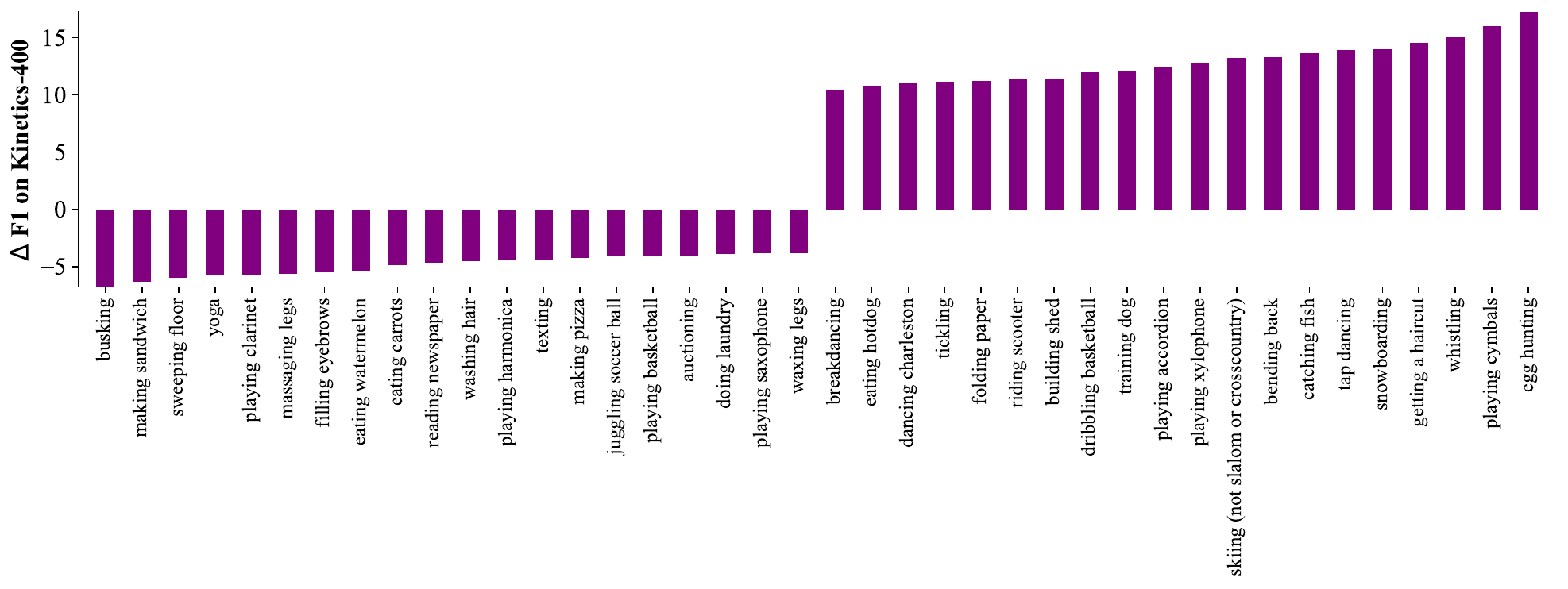}
    \vspace{-0.1in}
    \captionof{figure}{\textbf{Gain of \OURS over baseline on Action Recognition (per class).} We plot the gain in per-class F1-score on the \kineticsShort dataset for the top twenty and bottom twenty classes.
    The baseline model is first pretrained on \imnet and then fine-tuned on \kineticsShort whereas \OURS is trained jointly on \imnet, \kineticsShort and the single-view 3D \sunrgbd dataset.
    \OURS improves the F1 score on 308 out of the 400 total classes.
    }
    \label{fig:per_class_gain_k400}
\end{figure*}

In terms of preprocessing, at train time we sample a 32 frame video clip at stride 2 from the full video using temporal segment sampling as in~\cite{liu2021video}. We scale the short side of the video to 256px, take a 224px random resized crop, followed by RandAugment and Random Erasing.
At test time, we again sample a 32 frame clip with stride 2, scale the short side to 224px and take 3 spatial crops along the longer axis to get $224 \times 224$ crops. The final predictions are averaged over these crops.

For comparison to the \sota in~\cref{tab:sota-finetune-video-cls}, when finetuning \OURS models trained with \imnetFullShort, we found slightly different hyperparameters to perform better.
For \sthsth, we used peak learning rate of $1.2\cdot{10}^{-3}$ over 150 epochs.
For \epic, we used weight decay of 0.004, over 100 epochs, peak learning rate of $4\cdot{10}^{-4}$, with the same learning rate schedule for backbone and head. We also used cutmix augmentation and label smoothing.
All other hyperparameters in both cases were as described earlier. We also use EMA with similar settings as used during pretraining.

\subsection{Single-view 3D Tasks}
\label{app:details_transfer_depth}
\par \noindent \textbf{NYU Scene classification.}
We follow the setup from~\cite{gupta2013perceptual} for scene classification and use 10 classes derived from the original 19 classes. In~\cref{tab:sota-finetune-nyu} (classification) the best Swin B and Swin L models were trained for 200 epochs with starting learning rate of $5\times10^{-3}$, weight decay of 0 for Swin B and $1\times10^{-4}$ for Swin L. All other hyperparameters were as described earlier.

\par \noindent \textbf{NYU RGBD Segmentation.} We follow the training and evaluation setup from~\cite{cao2021shapeconv}.
We follow the Swin segmentation architecture which uses an UperNet~\cite{xiao2018unified} head with the Swin trunk.
All models are finetuned with AdamW~\cite{loshchilov2017decoupled} with a weight decay of $0.01$.
The learning rate follows a Polynomial Decay (power 1) schedule and starts at $0.00006$.
We warmup the learning rate for 1500 iterations and train the model with a batchsize of 32.
All the depth maps in NYU are converted into disparity maps by using the camera baseline and focal length of the Kinect sensor.

\subsection{$k$-NN experiments}

\paragraph{Extracting depth on \imnet.}
We ran a monocular depth-prediction model~\cite{ranftl2020towards} on the \imnetShort train set. We used the pretrained \texttt{dpt\_large} model and followed the input image preprocessing steps as provided in~\cite{ranftl2020towards}.

\paragraph{Classifying \imnet using different modalities.}
For the experiments involving classification using different modalities, we extract features from the \imnetShort train set using the RGB, RGBD or just Depth (D) modalities, and on \imnetShort validation set using the RGB modality.
We follow the $k$-NN protocol from~\cite{caron2021emerging} for evaluation and briefly describe it next.
We extract the stage 3~\cite{liu2021swin} features and $L_2$ normalize them. For each validation feature as the query, we retrieve the nearest neighbors from the train set using euclidean distance, and take the top-$k$ closest matches. For each match we create a one-hot vector using its ground truth label, and scale it by $e^{s/\tau}$, where $s$ is the dot product between the feature of the matched image the query image, and $\tau$ is a temperature hyperparameter (set to 0.07). We compute an effective prediction for the query by summing the top-$k$ one-hot vectors.
Similar processing is used for the visualizations in~\cref{fig:teaser} and~\cref{fig:knn_rgb_depth}.

\section{Other Results}
\label{app:other_results}

\noindent \textbf{Results on UCF-101.}
We also evaluate \OURS on another popular (albeit smaller) video recognition benchmark, UCF-101~\cite{soomro2012ucf101}. As shown in~\cref{tab:ucf101}, \OURS pre-training is effective for sports action recognition in UCF-101 as well.
Note that the results shown are with RGB modality only; the state-of-the-art on these datasets often leverages additional features such as optical flow, dense trajectories (IDT) etc.

\begin{table}
    \centering
    \setlength{\tabcolsep}{3pt}
    \begin{tabular}{c|ccH}
        & VideoSwin-B & \OURS (Swin-B) & \OURS (w/ \imnetFullShort) \\
        \hline
        3-split accuracy & 96.9 %
        & \bf 98.2 %
        & 98.0 \\  %
    \end{tabular}
    \caption{{\bf UCF-101.} As in~\cref{tab:inflation_vs_us_finetuned}, the VideoSwin model is inflated from \imnetShort and pre-trained on \kineticsShort. \OURS is pre-trained with \imnetShort, \kineticsShort and \sunrgbd. Both models are then finetuned and evaluated on UCF-101 for each split separately. Performance reported is averaged over the standard 3 splits. }\label{tab:ucf101}
\end{table}

{\noindent \bf Low-data regime fine-tuning.}
We analyzed low-shot versions of the \placesThree benchmark (models from~\cref{tab:inflation_vs_us_finetuned}).
As shown in~\cref{tab:low-shot}, \OURS outperforms the modality-specific baseline in the low-shot regime too.

\begin{table}
\centering
\setlength{\tabcolsep}{3pt}
\begin{tabular}{l|ccccHHHH}
  \bf Method & \multicolumn{4}{c}{\bf \placesThree} \\
  & 1\% & 2\% & 5\% & 10\% & 5\% & 10\% & 20\% & 50\% \\
  \hline
  \OURS & \bf 46.2 & \bf 49.0 & \bf 51.5 & \bf 53.9 & 2.2 & \bf 3.8 & \bf 11.0 & \bf 28.1 \\
  Image-specific & 44.8 & 47.9 & 50.9 & 53.4 & 2.2 & 3.3 & 9.6 & 25.7 \\
\end{tabular} %
\caption{{\bf Low-shot finetuning.} Performance of finetuning \OURS on low-shot versions of the \placesThree dataset.}\label{tab:low-shot}
\end{table}

{\noindent \bf Per-class gains.} We present the gain of \OURS over the VideoSwin baseline (\cref{sec:in1k_comparisons} of the main paper) in~\cref{fig:per_group_gain_k400_full,fig:per_class_gain_k400}.

\end{document}